\documentclass[preprint,12pt,numbers]{elsarticle}

\usepackage{amssymb}
\usepackage{graphicx}
\usepackage{amsmath}
\usepackage{multirow} 
\usepackage{xcolor}
\usepackage{booktabs}
\usepackage{pifont}
\usepackage{subcaption}
\usepackage[table]{xcolor}
\usepackage{hyperref}
\usepackage[sort&compress]{natbib}

\journal{Computers in Biology and Medicine}
\setcitestyle{square,numbers,sort&compress}
\begin{document}

\begin{frontmatter}



\title{Semi-Supervised Multi-Task Learning for Interpretable Quality Assessment of Fundus Images}



\author[inst1]{Lucas~Gabriel~Telesco}

\author[inst2]{Danila~Nejamkin}
\author[inst2]{Estefan\'ia~Mata}
\author[inst2]{Francisco~Filizzola}
\author[inst2]{Kevin~Wignall}
\author[inst2]{Luc\'ia~Franco~Troilo}
\author[inst2]{Mar\'ia~de~los~\'Angeles Cenoz}
\author[inst2]{Melissa~Thompson}
\author[inst2]{Mercedes~Legu\'ia}

\author[inst1]{Ignacio~Larrabide}
\author[inst1]{Jos\'e~Ignacio~Orlando}

\affiliation[inst1]{organization={UNCPBA, CONICET, Yatiris Group, Pladema Institute},
            addressline={Campus Universitario}, 
            city={Tandil},
            postcode={7000}, 
            state={Buenos Aires},
            country={Argentina}}

\affiliation[inst2]{organization={Servicio de Oftalmología, Hospital de Alta Complejidad en Red ``El Cruce'' Dr.~Néstor~Carlos~Kirchner},
,
            addressline={Av. Calchaquí 5401}, 
            city={Florencio Varela},
            postcode={1888}, 
            state={Buenos Aires},
            country={Argentina}}

\begin{abstract}
Retinal image quality assessment (RIQA) supports computer-aided diagnosis of eye diseases. However, most tools classify only overall image quality, without indicating acquisition defects to guide recapture. This gap is mainly due to the high cost of detailed annotations. In this paper, we aim to mitigate this limitation by introducing a hybrid semi-supervised learning approach that combines manual labels for overall quality with pseudo-labels of quality details within a multi-task framework. Our objective is to obtain more interpretable RIQA models without requiring extensive manual labeling. Pseudo-labels are generated by a Teacher model trained on a small dataset and then used to fine-tune a pre-trained model in a multi-task setting. Using a ResNet-18 backbone, we show that these weak annotations improve quality assessment over single-task baselines (F1: 0.875 vs. 0.863 on EyeQ, and 0.778 vs. 0.763 on DeepDRiD), matching or surpassing existing methods. The multi-task model achieved performance statistically comparable to the Teacher for most detail prediction tasks ($p > 0.05$). In a newly annotated EyeQ subset released with this paper, our model performed similarly to experts, suggesting that pseudo-label noise aligns with expert variability. Our main finding is that the proposed semi-supervised approach not only improves overall quality assessment but also provides interpretable feedback on capture conditions (illumination, clarity, contrast). This enhances interpretability at no extra manual labeling cost and offers clinically actionable outputs to guide image recapture.
\end{abstract}

\begin{graphicalabstract}
\includegraphics[width=\linewidth]{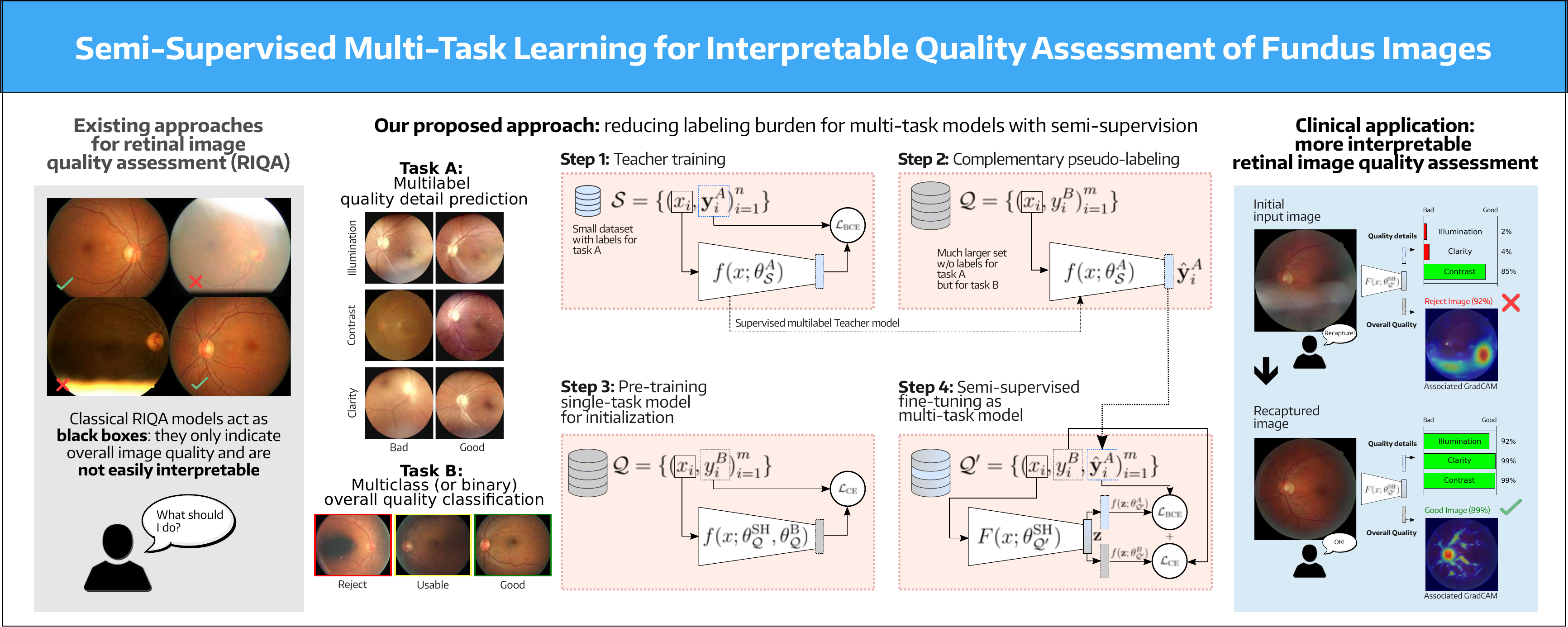}
\label{fig:graphical-abstract}
\end{graphicalabstract}


\begin{highlights}
    \item Apply semi-supervised learning to reduce label costs for RIQA multi-task models
    \item Teacher model generates pseudo-labels to enhance interpretability in RIQA  
    \item Pseudo-labels’ noise aligns with expert variability in detail quality labeling
    \item Multi-task model improves RIQA performance over single-task approaches  
    \item We publicly release new accurate EyeQ quality detail labels for future research
\end{highlights}


\end{frontmatter}


\section{Introduction}
\label{sec:introduction}

The effectiveness of fundus image–based automated systems for disease screening and diagnosis heavily depends on input image quality \cite{raj2019fundus,cao2023application,lin2020retinal}. 
Accordingly, these platforms often include computer-aided retinal image quality assessment (RIQA) tools trained to discriminate suboptimal fundus images from those suitable for downstream analysis \cite{guo2023learning,abdel2018performance,costa2017eyequal}.
Most existing algorithms rely on supervised learning and either perform structural analysis using anatomical segmentations \cite{zhou2020retinal,xu2023deep,shen2020modeling}, or characterize images holistically using global features \cite{wang2015human,raj2020multivariate} or deep neural networks \cite{fu2019evaluation,leonardo2022impact,muddamsetty2021multi}. 

Typically, these solutions classify overall quality as either good vs. bad \cite{li2022deep,yu2017image,zhou2018fundus,liu2022degradation}, rejectable/usable/good \cite{fu2019evaluation,qayyum2022single,laurik2022assessment}, or even finer grades~\cite{muddamsetty2021multi}.
While such labels help exclude low-quality images from screening, they usually lack interpretability: they do not indicate acquisition mistakes, which technicians need to correct captures before patients leave \cite{shen2020domain}. Classical explainability methods such as class activation maps \cite{shen2020domain,xu2022dark,abramovich2023fundusq} offer partial visual cues, but they may fail to explicitly highlight concrete problems (poor illumination, defocus, low contrast) because they depend on both output accuracy and the model’s internal patterns.

Incorporating complementary information into training--e.g., in multilabel \cite{konig2024quality} or multi-task learning settings \cite{shen2018multi,shen2020domain}--can improve quality classification by steering attention to imaging characteristics and by exposing additional, inherently interpretable outputs.
However, this comes at the cost of extra manual annotations \cite{jin2023mshf}. As a result, models are often trained on smaller datasets, limiting coverage of imaging conditions and hampering generalization across tasks \cite{raj2020multivariate}.

Accurate and interpretable RIQA is therefore needed at the point of acquisition, yet detailed quality annotations (illumination, clarity, contrast) are costly and scarce. Our work is motivated then by two goals: (i) embed intrinsic interpretability into RIQA without a significant impact in the annotation burden, and (ii) deliver actionable feedback that enables technicians to correct capture settings in real time, improving clinical workflow and patient care. 

We propose to address this gap by means of a semi-supervised multi-task approach. In particular, we train a model with ground-truth overall quality labels and pseudo-labels for imaging conditions (e.g., illumination, clarity, contrast) obtained from a Teacher model trained on a small labeled dataset. Our hypothesis is that the noise introduced by pseudo-labeling these conditions is comparable to the intrinsic variability among professionals performing this task. Thus, we can still benefit from task complementarity to obtain more comprehensive, interpretable outputs at a fraction of the original labeling cost. We further hypothesize that this auxiliary supervision helps the overall-quality branch attend to patterns tied to acquisition conditions, improving discrimination performance and the resulting GradCAMs.

\begin{figure}[t]
\centering
\includegraphics[width=0.9\textwidth]{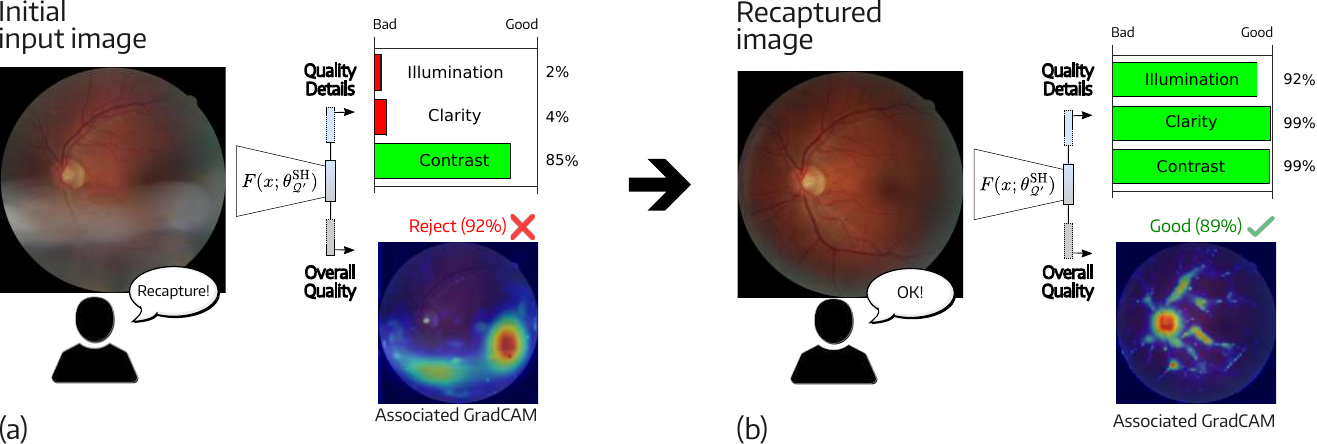}
\caption{Example use case of the proposed semi-supervised multi-task RIQA, showing the assessment of an initial low-quality image (a) and its improved, recaptured version (b).}
\label{fig:principal}
\end{figure}

We validate these hypotheses by training a multilabel Teacher on MSHF~\cite{jin2023mshf} and then evaluating our multi-task model on EyeQ~\cite{fu2019evaluation} and DeepDRiD~\cite{liu2022deepDRID}. Using a simple architecture, our approach attains overall-quality results that match or surpass state-of-the-art single-task baselines, while producing quality-detail predictions that are statistically comparable to the Teacher model. The method also yields more faithful class activation maps, helping users verify the outputs. Figure \ref{fig:principal} depicts an exemplary use case of the proposed approach. For a low-quality image, the model indicates both its overall quality and which acquisition conditions to adjust (in the example, illumination and clarity), while also highlighting problematic regions with a GradCAM. This could be used by a technician to guide the reacquisition process, improving the quality of the scan.

The remainder of this paper is organized as follows. Sections~\ref{subsec:introduction-related-works} and \ref{subsec:introduction-contributions} summarize the state of the art and our contributions, respectively. Section~\ref{sec:methods} details the proposed approach, and Section~\ref{sec:experimental-setup} presents the experimental setup, including materials, evaluation metrics, implementation, and baselines. Section~\ref{sec:results} reports results, Section~\ref{sec:discussion} discusses them, and Section~\ref{sec:conclusions} concludes the paper.

\subsection{Related works}
\label{subsec:introduction-related-works}

\subsubsection{Interpretable retinal image quality assessment}
\label{subsubsec:introduction-related-works-explainability}

Automated retinal image quality assessment (RIQA) aims to identify images that are adequate for subsequent diagnostic tasks, distinguishing them from those that need to be recaptured or discarded \cite{raj2019fundus}. Most methods either assess the visibility of key anatomical landmarks \cite{zhou2020retinal,xu2023deep} or characterize global appearance \cite{wang2015human,raj2020multivariate}, using manually engineered features \cite{fasih2014retinal, dias2014retinal, csevik2014identification} or deep models that learn features directly from data \cite{fu2019evaluation,leonardo2022impact,muddamsetty2021multi}. Table \ref{tab:qualitative_comparison_full} summarizes some key characteristics of the most recent approaches, including their quality outputs, training labels, if they implement any label-efficiency strategy, and if they considered any specific approach for additional interpretability. Older methods are not included in the table but explained in the sequel.

\begin{table}[t!]
\centering
\caption{Comparison of key features in recent RIQA methods.}
\label{tab:qualitative_comparison_full}
\resizebox{\textwidth}{!}{
\begin{tabular}{lcccc}
\toprule
\textbf{Method} & \textbf{Quality output} & \textbf{Training labels} & \textbf{Label-efficiency strategy} & \textbf{Interpretability} \\
\midrule
Muddamsetty \& Moeslund 2021 \cite{muddamsetty2021multi} & Only overall & Image-level & Supervised & None \\
Leonardo et al. 2022 \cite{leonardo2022impact}   & Only overall & Image-level & Supervised & None \\
Abramovich et al. 2023 \cite{Abramovich2022FundusQNetAR} & Only overall & Image-level & Semi-supervised & CAM-only \\
Engelmann et al. 2023 \cite{engelmann2023quickqual} & Only overall & Image-level & Lightweight head & None \\
Xu et al. 2023 \cite{xu2023deep} & Only overall & Image-level & Supervised & Priors \\
Yi et al. 2023 \cite{yi2023label} & Overall + details & None & Zero-shot & None \\
Huang et al. 2024 \cite{huang2024enhancing} & Only overall & Image-level & Supervised & CAM-only \\
König et al. 2024 \cite{konig2024quality} & Overall + details & Image-level & Supervised & Additional task \\
Guo et al. 2024 \cite{Guo2024RefinedIQ} & Overall + details & Image-level & Supervised & Additional task \\
\midrule
\textbf{Proposed approach} & \textbf{Overall + details} & \textbf{Image-level} & \textbf{Semi-supervised} & \textbf{Additional task + CAM} \\
\bottomrule
\end{tabular}
}
\end{table}

Segmentation-based approaches assume that landmarks such as the optic disc \cite{kohler2013automatic, bhatkalkar2020automated}, blood vessels \cite{nugroho2014contrast, csevik2014identification}, and/or the fovea \cite{shao2017automated, csevik2014identification} should be clearly visible in any good-quality image \cite{fleming2006automated, paulus2010automated}. They use pre-trained segmentation models to identify regions of interest and then evaluate those masks to decide overall quality \cite{nugroho2014contrast, shao2017automated, csevik2014identification}. Classical classifiers such as Support Vector Machines (SVMs) \cite{csevik2014identification} are combined with hand-crafted descriptors (e.g., vessel continuity and thickness \cite{wen2007automated}) in this case. More recent methods apply deep networks directly to segmentation masks, reducing manual feature design \cite{bhatkalkar2020automated, saha2017deep, zhou2020retinal, xu2023deep}. These pipelines are inherently interpretable, since segmentations are self-explanatory, but they depend on accurate pre-trained models that require costly pixel-level annotations \cite{lin2020retinal, costa2017eyequal, lyu2022fractal}. Xu et al. \cite{xu2023deep} overcomes this limitation by using a pure unsupervised image-processing based approaches for segmenting  anatomical structures.

Feature-based approaches, on the other hand, apply designed filters to detect acquisition issues such as blur, poor contrast, or suboptimal illumination \cite{shao2017automated, raj2019fundus, davis2009vision}. These features are then combined with SVMs \cite{csevik2014identification, wang2015human, fasih2014retinal} or decision trees \cite{wang2015human, shao2017automated} for classification. While they avoid the need for segmentation, hand-crafted filters can generalize poorly and their abstractions are less interpretable to clinicians \cite{raj2019fundus, gonccalves2023image}.

Deep learning is nowadays the standard for RIQA. Most methods rely on convolutional neural networks (CNNs) \cite{you2019fundus, kim2017deep,fu2019evaluation, raj2019fundus}, and recent work explores Transformer-based architectures \cite{huang2024enhancing} as an alternative. Advances include tailored input representations \cite{fu2019evaluation, karlsson2021automatic}, additional inputs~\cite{xu2023deep}, and architectures \cite{fu2019evaluation,raj2019fundus,gonccalves2023image,huang2024enhancing}. Fu et al. \cite{fu2019evaluation}, for example, introduced a multi-input CNN that processes three color spaces to predict good/acceptable/ungradable quality. Guo et al. \cite{guo2023learning} proposed a dual-channel attention CNN with cost-aware regularization and label smoothing. Recent work has also varied task design and losses. Muddamsetty \& Moeslund \cite{muddamsetty2021multi}, for example, studied multi-level quality classification, highlighting performance drops under class imbalance. Leonardo et al. \cite{leonardo2022impact}, on the other hand, used a regression loss for prediction. Finally, Xu et al. \cite{xu2023deep} employed segmentation masks as input priors for the CNN-based classification model.

Hybrid approaches have emerged to combine strengths across categories \cite{bhatkalkar2020automated, wang2021deep, wen2007automated, paulus2010automated, mahapatra2016retinal, abdel2016retinal}. Yu et al. \cite{yu2017image}, for instance, fused saliency-based and CNN features, classifying the joint embedding with an SVM. Shi et al. \cite{shi2022assessment}, on the other hand, distinguished good versus poor images and attributed poor quality to ocular abnormalities or image noise using hand-crafted features and a CNN with an SVM head. Engelmann et al. \cite{engelmann2023quickqual} recently showed that off-the-shelf features from a CNN plus a lightweight head can be effective without domain-specific pretraining, requiring then less training samples.

CNN-based methods deliver strong performance but often offer limited guidance on which capture factors degrade quality. Most models predict only overall quality (e.g., good/bad or rejectable/usable/good \cite{fu2019evaluation, raj2019fundus}), without featuring any explanatory mechanism. Recent literature explored mitigating this factor by modifying the formulations, i.e. injecting salient anatomical priors \cite{xu2023deep}. Others explore label-free, zero-shot strategies using language–vision features \cite{yi2023label}, which, however, may overvalue intact regions in partially degraded images. Multi-label formulations have started to address this gap by predicting both overall quality and capture-related issues (focus, contrast, illumination) \cite{wang2020multi, shen2020domain, konig2024quality}. K{\"o}nig et al. \cite{konig2024quality}, for example, use a multi-label network to report specific defects at the expense of additional labeling.

Explainability methods for CNNs typically aim to generate approximate maps highlighting regions of the input image that were taken into account by the model for decision-making \cite{samek2019explainable}. Their main goal is to help experts verify whether the model has focused on the correct features to make predictions or not \cite{cen2021automatic}. In RIQA, these maps should correlate with areas of poor imaging and can help technicians understand acquisition mistakes. In medical imaging \cite{zhao2023multi}, one of the most widely used explainability techniques are GradCAMs \cite{zhou2016learning, selvaraju2017grad}, which correspond to heatmaps obtained by calculating gradients of the target concept with respect to the final convolutional layer of a CNN \cite{selvaraju2017grad}. GradCAMs have been applied in RIQA models to explain classification outputs \cite{manne2023diagnostic, abramovich2023fundusq, ubukata2024fundus}. In this paper we demonstrate that pursuing a multi-task approach in a semi-supervised learning fashion can result in better overall quality predictions. Producing outputs that are more accurate enables GradCAMs produced for the most probable class to be better aligned with image observations, compared to a mistaken output from the single-task counterpart.

Recent CNN- and Transformer-based RIQA systems achieve high accuracy and sometimes provide saliency maps, but most predict only an overall grade and depend on task-specific labels to expose capture defects. Multi-label models increase actionability (focus/contrast/illumination) but raise annotation costs and can be brittle under class imbalance. Lightweight and zero-shot variants cut labeling yet trade off granularity or stability on localized artifacts. These gaps motivate a label-efficient multi-task design that predicts overall quality and capture conditions jointly and supports faithful GradCAM explanations.

\subsubsection{Multi-task and semi-supervised learning}
\label{subsubsec:introduction-related-works-multitask}

Multi-task learning is a subset of machine learning in which a single model is trained to solve multiple related tasks simultaneously \cite{kendall2018multi}. By leveraging shared information across tasks, these models can outperform models trained in isolation \cite{ruder2017overview, Zhang2018AnOO, zhang2021survey}. A common implementation uses architectures with multiple output branches connected to a shared backbone, enabling the model to exploit common patterns among tasks. This approach has been successfully applied in natural language processing \cite{chen2024multi}, speech recognition \cite{dai2021weakly}, and computer vision \cite{kendall2018multi,Vandenhende2022multitaskLF}. 

In retinal image analysis, multi-task learning has been used for tasks such as the simultaneous diagnosis of diabetic retinopathy (DR) and diabetic macular edema \cite{li2019canet}, lesion segmentation \cite{zhou2020benchmark}, and blood vessel segmentation \cite{wang2020multi}, among others. In the context of RIQA, research has focused on exploiting auxiliary tasks to boost accuracy while providing more interpretable feedback. Shen et al. \cite{shen2020domain}, for instance, proposed a multi-task deep learning framework that assesses overall image gradability while simultaneously evaluating specific quality factors, including artifacts, clarity, and field definition, and integrates anatomical landmark detection as an extra task to enhance assessment accuracy. Similarly, Guo et al. \cite{Guo2024RefinedIQ} introduced a multi-task model that evaluates fundus image quality based on a three-level criterion for key subcategories (location, clarity, and artifacts), incorporating auxiliary tasks such as optic nerve head and macula localization, as well as field-of-view classification. These advances improve both the accuracy and interpretability of RIQA, which is essential for reliable ocular disease diagnosis \cite{Guo2024RefinedIQ, Abramovich2022FundusQNetAR, shen2020domain}.

The primary limitation of multi-task learning is the need for additional labels for each added task, which increases the expert annotation burden \cite{dai2021weakly, karimi2020deep, algan2021image}. As with any supervised task, these annotations are subject to inter- and intra-observer variability, which may worsen as more tasks are integrated \cite{lemay2022label, tanno2019learning}.

Semi-supervised learning is a promising alternative to mitigate this limitation~\cite{Wang2021SemisupervisedML}. In this paradigm, large unlabeled datasets are combined with smaller labeled sets to enhance accuracy and generalization while reducing annotation costs \cite{huynh2022semi}. This approach has recently gained popularity in retinal imaging applications, e.g., DR classification~\cite{lecouat2018semi}, fundus image representation learning~\cite{yap2021semi}, and optic disc segmentation \cite{Meng2024MultigranularityLO}.

To leverage unlabeled samples, several techniques have been developed, with Noisy Student being one of the most widely adopted \cite{xie2020self}. In this method, a Teacher model is first trained on a small set of annotated images and then used to generate pseudo-labels for the unlabeled data. These pseudo-labeled samples, combined with the original labeled data, are used to train a Student network, enhancing overall performance \cite{moris2024semi}.

This framework can be exploited in a multi-task setting, for example by training a single-task model on a small labeled subset to pseudo-label a larger unlabeled set and then combining these labels with manual annotations to improve multiple tasks jointly \cite{dai2021weakly}. This strategy has been only partially studied for retinal image analysis in general \cite{liu2023pldmlt} and, to our knowledge, has not yet been explored in RIQA. To the best of our knowledge, the closest existing approach is that proposed by Dai et al. \cite{dai2021weakly}, which combines pseudo-labels and manual labels from different tasks within the same training scheme for speech recognition; and \cite{liu2023pldmlt}, which leverages pseudo-labels for DR grading together with manually generated pixel-level lesion masks to improve lesion segmentation.

Inspired by the application of these hybrid approaches in other domains, our work introduces a semi-supervised, multi-task learning framework for interpretable quality assessment. This formulation consists of training a single model to solve two tasks simultaneously: overall quality classification using manual labels, and prediction of specific capture details using pseudo-labels. These are generated by a Teacher model previously trained on a small dataset, which substantially reduces the annotation burden for the auxiliary task on larger datasets.

\subsection{Contributions}
\label{subsec:introduction-contributions}

Our contributions are fourfold: (i) We propose a semi-supervised scheme for RIQA multi-task learning that reduces the need for extra annotations by using pseudo-labels for the auxiliary task. A Teacher model trained on a small expert-annotated set produces these pseudo-labels, which we use to fine-tune a network initially pre-trained only for overall quality assessment; (ii) We show that pseudo-label noise is comparable to inter-observer variability. Under this regime, the multi-task model improves the primary task and statistically matches expert performance on the auxiliary task, while requiring far fewer manual labels and offering greater interpretability than its single-task counterpart; (iii) We analyze how multi-task learning guides the primary task toward correct predictions and yields more informative GradCAMs; (iv) We release expert annotations of capture conditions for a subset of EyeQ images to support further research.

\section{Methods}
\label{sec:methods}

\begin{figure}[t]
\centering
\includegraphics[width=0.9\textwidth]{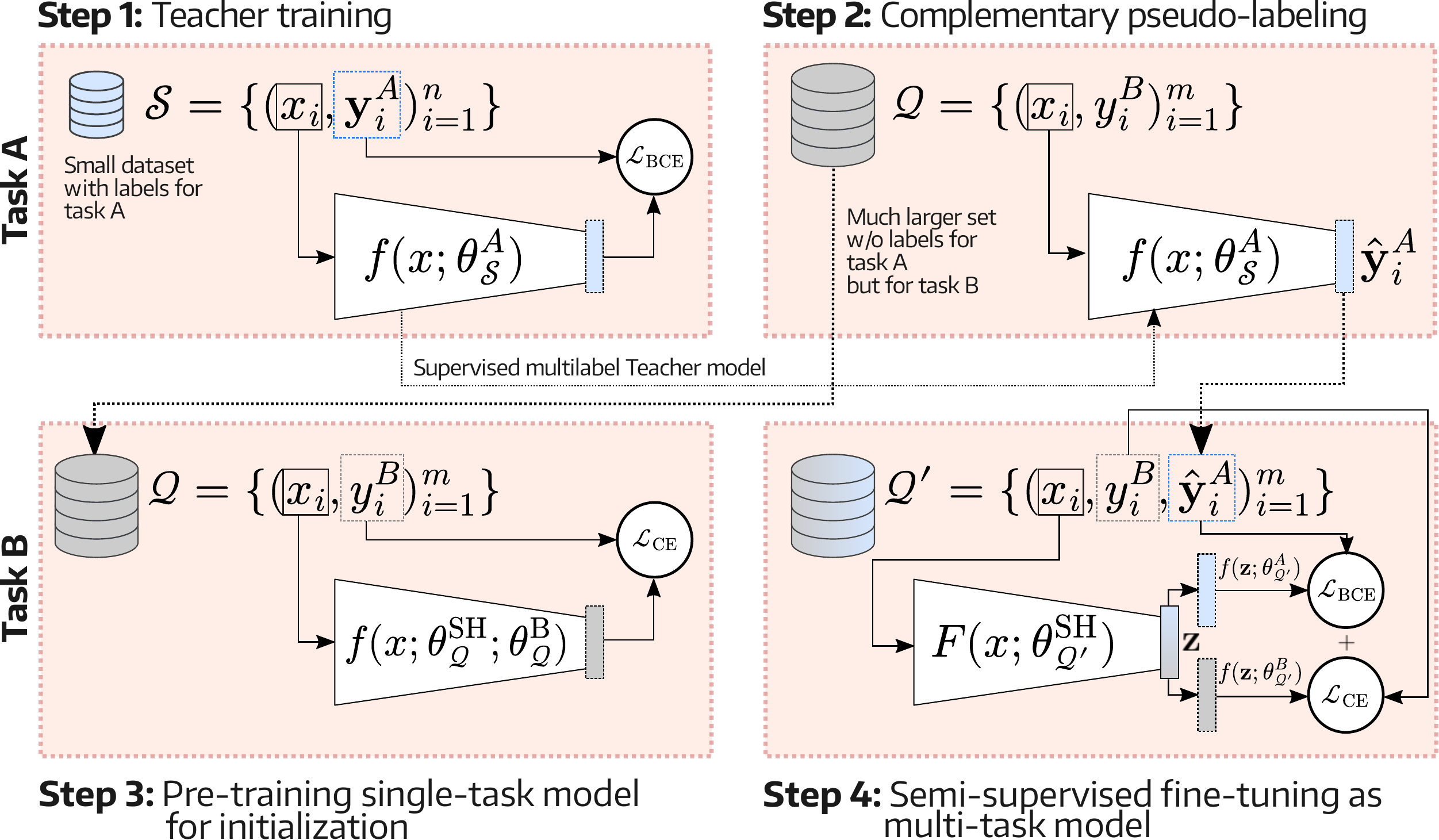}
\caption{Schematic representation of our proposed semi-supervised multi-task learning approach.}
\label{fig:schematic}
\end{figure}

We hypothesize that accurate multi-task RIQA models can be trained by combining manual annotations of overall image quality (task $B$) with pseudo-labels for imaging conditions (task $A$). Such models are expected to improve performance on task $B$ while increasing interpretability via the additional outputs for task $A$. Figure \ref{fig:schematic} schematizes the approach. First, a Teacher model is trained for task $A$ on a small, expert-annotated dataset of imaging conditions (i.e., illumination, clarity, and contrast labeled as good/bad). This Teacher then generates pseudo-labels on a larger dataset in which imaging conditions are unknown but overall image quality is manually annotated (Section \ref{subsec:methods-Teacher-A}). Next, we pre-train a single-task model for task $B$ using these ground-truth quality labels and then adapt it by adding an auxiliary prediction branch for task $A$. Finally, we fine-tune the resulting multi-task model with pseudo-labels for task $A$ and manual labels for task $B$ (Section \ref{subsec:methods-student-multi-task}).

\subsection{Pseudo-labeling with a pre-trained multilabel Teacher model}
\label{subsec:methods-Teacher-A}

Let $\mathcal{S} = \{(x_i, \mathbf{y}_{i}^{A})_{i=1}^n\}$ be a small dataset of $n$ fundus images $x_i \in \mathcal{X}$ with corresponding labels $\mathbf{y}_{i}^{A}$ for task $A$, where $\mathcal{X}$ denotes the set of all possible retinographies. Task $A$ is multilabel: each image $x_i$ maps to a binary vector of quality-detail annotations $\mathbf{y}_{i}^{A} \in \mathcal{Y}^A = \{0,1\}^k$, where $k$ is the number of quality details (in our experiments, $k=3$ for illumination, clarity, and contrast labeled as good (1) or bad (0)). A straightforward approach, following \cite{konig2024quality}, is to train a multilabel classifier $f(x; \theta^{A}_\mathcal{S}): \mathcal{X} \to \mathcal{Y}^A$, where $f$ denotes the network and $\theta^{A}_\mathcal{S}$ are the parameters learned from $\mathcal{S}$ for task $A$. This training (step 1 in Figure~\ref{fig:schematic}) minimizes the binary cross-entropy loss $\mathcal{L}_\text{BCE}$ between ground-truth targets $\mathbf{y}^{A}$ and the model outputs $f(x; \theta^{A}_\mathcal{S})$:
\begin{equation}
\mathcal{L}_\text{BCE}  = -\frac{1}{n} \sum_{i=1}^n \sum_{j=1}^k \left( y_{i,j}^{A} \log f_j(x_i; \theta^{A}_\mathcal{S}) + (1 - y_{i,j}^{A}) \log (1 - f_j(x_i; \theta^{A}_\mathcal{S})) \right),
\end{equation}
where $y^{A}_{i,j}$ is the ground-truth label for the $j$-th quality detail of image $x_i$, and $f_j(x_i; \theta^{A}_\mathcal{S})$ is the predicted probability for that detail.

Because $\mathcal{S}$ is small, the resulting model may overfit and generalize poorly. Nevertheless, its predictions can provide a useful--albeit noisy--training signal for a related task such as overall quality classification within a multi-task setting (see Section~\ref{subsec:methods-student-multi-task}). Formally, let $\mathcal{Q} = \{(x_i, y_i^{B})_{i=1}^m\}$ be a larger dataset ($m \gg n$) of unseen fundus images with manual labels for task $B$. In our setup, task $B$ is overall quality classification, with labels $y_i^{B} \in \mathcal{Y}^{B}$ whose categories depend on the dataset: for EyeQ~\cite{fu2019evaluation}, labels correspond to rejectable, acceptable, and good image quality ($|\mathcal{Y}^{B}| = 3$), whereas for DeepDRiD~\cite{liu2022deepDRID} they indicate bad and good image quality ($|\mathcal{Y}^{B}| = 2$).

We use $f(x; \theta^{A}_\mathcal{S})$ as the Teacher to generate pseudo-labels for the quality details of images in $\mathcal{Q}$ (step 2 in Figure~\ref{fig:schematic}). This produces a label-augmented set $\mathcal{Q}' = \{(x_i, y_i^{B}, \hat{\mathbf{y}}_i^{A})_{i=1}^m\}$, where $\hat{\mathbf{y}}_i^{A} = f(x_i ; \theta^{A}_\mathcal{S})$. We later use $\mathcal{Q}'$ to fine-tune the multi-task model.

\subsection{Multi-task Student for interpretable image quality assessment}
\label{subsec:methods-student-multi-task}

Multi-task learning leverages shared information among related tasks, improving performance relative to training them independently \cite{kendall2018multi}. One common implementation is hard parameter sharing, where a single model solves multiple tasks with shared parameters and task-specific heads. Following the notation in \cite{ju2021synergic}, we formalize such a model as a sequence of forward passes:
\begin{equation}
\left\{ \begin{array}{l}
            \mathbf{z} = F(x_i; \theta^\text{SH}) \\
            f_A = F(\mathbf{z}; \theta^{A}) \\
            f_B = F(\mathbf{z}; \theta^{B})
        \end{array} \right.
\end{equation}
where $\theta^\text{SH}$ are shared parameters, $\theta^{A}$ and $\theta^{B}$ are task-specific, and $\mathbf{z}$ is the common feature representation.

We train a single multi-task Student with hard parameter sharing to solve $A$ (multilabel quality-detail prediction) and $B$ (multiclass or binary overall quality classification) simultaneously, using the pseudo-labels in $\mathcal{Q}'$ as targets for task $A$. Our assumption is that this setup encourages a shared representation $\mathbf{z}$ that captures quality-related details, improving performance on the overall quality task while increasing interpretability via the auxiliary outputs.

For pre-training, we optimize the multiclass cross-entropy loss $\mathcal{L}_\text{CE}$:
\begin{equation}
\mathcal{L}_\text{CE} = -\frac{1}{m} \sum_{i=1}^m \sum_{c=1}^C y_{i,c}^{B} \log \big( f_{B,c}(x_i; \theta^\text{SH}, \theta^{B}) \big),
\end{equation}
where $m$ is the number of samples in $\mathcal{Q}$, $C$ is the number of quality classes ($C\in\{2,3\}$ in our experiments), $y_{i,c}^{B}$ is a one-hot indicator (1 if the correct class for sample $i$ is $c$, 0 otherwise), and $f_{B,c}(x_i; \theta^\text{SH}, \theta^{B})$ denotes the $c$-th component of $f_B$. We use this loss to pre-train $(\theta^\text{SH}, \theta^{B})$ for quality assessment (step 3 in Figure~\ref{fig:schematic}). We then add an auxiliary prediction branch for task $A$, with parameters $\theta^{A}$ randomly initialized, and fine-tune all parameters $\theta = \theta^\text{SH} \cup \theta^{A} \cup \theta^{B}$ by minimizing the multi-task objective:
\begin{equation}
\label{eq:multi-task-objective}
    \mathcal{L} = \lambda_A \, \mathcal{L}_\text{BCE}(\hat{\mathbf{y}}_i^{A}, f_A) + \lambda_B \, \mathcal{L}_\text{CE}(y_i^{B}, f_B),
\end{equation}
where $\lambda_A$ and $\lambda_B$ weight the contribution of each task (step 4 in Figure~\ref{fig:schematic}). This additional signal helps learn shared features that better solve overall quality classification while capturing patterns that support interpretability.

\section{Experimental setup}
\label{sec:experimental-setup}

We empirically evaluate the proposed approach by (i) measuring improvements on the primary task of overall quality classification, and (ii) assessing the effectiveness of the semi-supervised auxiliary task that predicts quality details. This section describes the datasets used for training and evaluation (Section \ref{subsec:materials}), details the model backbones and training settings for the multi-task models and single-task baselines (Section \ref{subsec:experimental-setup-model-baseline}), and specifies the evaluation metrics (Section \ref{subsec:evaluation-metrics}).

\subsection{Materials}\label{subsec:materials}

All experiments used public fundus-image datasets—MSHF~\cite{jin2023mshf}, EyeQ~\cite{fu2019evaluation}, and DeepDRiD~\cite{liu2022deepDRID}--applying the same field-of-view (FOV) cropping strategy as in~\cite{castilla2023resnet} to remove non-informative regions while preserving the original aspect ratio.

MSHF contains 802 retinal images with binary labels for illumination, clarity, and contrast (task $A$), each annotated as good (1) or bad (0). Images were captured using Kowa Nonmyd DR-XJU cameras (FOV 45 degrees, resolution 1924$\times$1556), TRC-NW8 cameras (FOV 50 degrees, resolution 1924$\times$1556), and DEC200 portable cameras (FOV 60 degrees, resolution 2560$\times$1960). Of these, 422 scans correspond to patients with DR, 52 to glaucomatous subjects, and 328 to healthy individuals. The dataset provides a single split for training (642 images) and testing (160 images); we randomly held out 65 training images (10\%) for validation. We used MSHF to train the Teacher for quality-detail prediction and to evaluate task $A$.

EyeQ is a well-established RIQA dataset introduced in~\cite{fu2019evaluation}. It comprises 28792 images collected with more than 40 fundus cameras in a diabetic retinopathy screening program, with manual labels for overall quality--``rejectable'' (class 0), ``usable'' (class 1), and ``good'' (class 2)--for task $B$. The dataset is split into training (12543 images) and test (16249 images); we additionally sampled 1254 training scans (10\%) for validation. We used EyeQ to train and evaluate our multi-task approach and the single-task baselines. Because task-$A$ labels are not publicly available for EyeQ, we randomly sampled 160 test images, stratified by overall-quality class, and asked eight experienced ophthalmologists to label them using a custom application. For each image, a subset of experts with an odd number of members annotated the quality details following the criteria used for MSHF~\cite{jin2023mshf}; final labels were assigned by majority vote. We refer to this subset as \emph{EyeQ-D} and use it to study inter-observer variability for task $A$ and as a secondary evaluation set. To facilitate future comparisons, we publicly release these labels at \href{https://github.com/ltelesco/Semi-Supervised-Multi-Task-Learning-for-Interpretable-Quality-Assessment-of-Fundus-Images}{https://github.com/ltelesco/Semi-Supervised-Multi-Task-Learning-for-Interpretable-Quality-Assessment-of-Fundus-Images}.

Finally, DeepDRiD provides 2000 fundus images for RIQA, acquired with non-mydriatic digital cameras with FOVs ranging from 45 to 60 degrees, centered on the macula or the optic disc. It is partitioned into training (1200 images), validation (400 images), and test (400 images) sets, and includes binary labels for overall quality: ``poor'' (0) and ``good'' (1). We use DeepDRiD to complement EyeQ in our evaluation and to study how label granularity and training-set size influence results.

\subsection{Model implementation and baselines}
\label{subsec:experimental-setup-model-baseline}

All experiments were carried out using ResNet-18~\cite{he2016deep} backbones initialized with ImageNet-pretrained weights and implemented in PyTorch 1.11. Deeper or more complex backbones, such as Vision Transformers~\cite{dosovitskiy2020image}, were not considered due to the risk of overfitting on our limited training sets. All models were trained for up to 115 epochs using SGD with momentum 0.9; the initial learning rate was 0.01 and was halved at epochs 30, 60, and 80. For final evaluation, we used only the best-performing checkpoint across epochs, selected by validation performance.

The Teacher for task $A$ (Teacher-A, also a ResNet-18) was fine-tuned for detailed quality prediction on MSHF using a multi-label classification head with sigmoid activations. No data augmentation was applied, consistent with common practice in Noisy Student training~\cite{xie2020self}. 

The multi-task model was first trained for overall quality classification (task $B$) using ground-truth labels from either EyeQ or DeepDRiD, depending on the experiment, and was then fine-tuned in a multi-task setting by adding a multi-label branch for task $A$. We refer to the resulting models as MT-EyeQ and MT-DeepDRiD, respectively. In all cases, the original backbone was a ResNet-18 architecture, that modified for multi-task prediction adding the auxiliary branch. For data augmentation, we used a strategy inspired by RandAugment~\cite{cubuk2020randaugment}, with transformations including rotations, horizontal and vertical flips, and small color perturbations to avoid severe degradations that could alter label semantics. Each training image underwent at most seven transformations, uniformly sampled with replacement, with random strengths chosen from predefined intervals. Model selection was based on F1-score for overall quality assessment on the validation set. Hyperparameters $\lambda_A$ and $\lambda_B$ in the multi-task objective were tuned using the same criterion.

Pre–fine-tuning multi-task models served as single-task baselines (ST-EyeQ and ST-DeepDRiD). To control for potential effects of training duration, we extended their training by an additional 115 epochs and retained the best checkpoint over all epochs for comparison. In all cases, the best-performing single-task models were those obtained before this extension.

\subsection{Evaluation metrics}\label{subsec:evaluation-metrics}

To address class imbalance, we evaluate predictions for overall quality and quality details using F1-score (F1), precision (Pr) and recall (Re). We also included accuracy (Acc) for reference purposes. For multiclass settings such as in EyeQ, we report macro F1, i.e., the unweighted mean of the one-vs-all per-class F1 scores. Results in DeepDRiD, on the other hand, are reported in a binary setting, considering the bad quality class as the positive one. When comparing to prior work, we either use the values reported in the original papers,  compute them from the authors’ published confusion matrices~\cite{muddamsetty2021multi,xu2023deep,yi2023label,Guo2024RefinedIQ}, or re-trained the model in new data~\cite{engelmann2023quickqual}. We assess statistical significance with one- or two-tailed Wilcoxon signed-rank tests and bootstrap resampling.

\section{Results}
\label{sec:results}

We report results from two perspectives, both quantitatively and qualitatively. First, we measure the effect of the semi-supervised auxiliary task on overall quality classification by comparing the multi-task model to its single-task counterpart and to established RIQA baselines (Section~\ref{subsec:results:quality}). Second, we assess the auxiliary task itself--predicting quality details--on MSHF and EyeQ-D, and compare its performance against the Teacher model and inter-observer variability (Section~\ref{subsec:results:capture}).

\subsection{Overall quality classification}
\label{subsec:results:quality}

To quantify the effect of adding the quality-detail prediction task, we evaluated the multi-task and single-task models for overall quality classification on EyeQ and DeepDRiD. Table~\ref{tab:comparativa} lists the results in terms of F1, Pr, Re, and Acc, comparing our proposed approach and its single-task baseline against several state-of-the-art methods. Both models were trained under identical conditions with the same convolutional backbone; the multi-task variant differs only by the auxiliary branch for quality details (see Section~\ref{subsec:experimental-setup-model-baseline}).

MT-EyeQ achieved higher scores across all metrics than its single-task counterpart, with statistically significant improvements in every case ($p<0.05$). MT-DeepDRiD showed a similar trend, reaching a significant gain in F1 ($p<0.05$); the remaining metrics increased without statistical significance, except for Pr, where the single-task model was slightly higher.

For context, Table~\ref{tab:comparativa} also lists state-of-the-art EyeQ results, including a DenseNet-like architecture~\cite{huang2017densely} and a Swin Transformer~\cite{liu2021swin} as reported in~\cite{huang2024enhancing} and several other recent approaches. Overall, MT-EyeQ is competitive with--or better than--recent methods, most of which provide limited interpretability and predict only overall quality (see Section~\ref{subsubsec:introduction-related-works-explainability}). The exception is the approach by K\"onig et al.~\cite{konig2024quality}, which predicts binary labels for overall quality and multiple acquisition details in a multilabel setup, then adjusts decision margins to produce three classes. However, their method yields lower F1, Pr, and Re than both ST-EyeQ and MT-EyeQ.

To produce a comparable evaluation on DeepDRiD, we re-trained QuickQual~\cite{engelmann2023quickqual} using these images. Results are also reported in Table~\ref{tab:comparativa}. Our multi-task model consistently outperformed the competing method, achieving higher F1, Pr, Re, and and Acc values.

\begin{table}[t!]
\caption{Results for overall quality classification on the EyeQ and DeepDRiD test sets.}
\centering
    \resizebox{0.99\textwidth}{!}{
    \begin{tabular}{cc|c|c|c|c}
    \hline
    \multicolumn{2}{c|}{\textbf{Model}}                                                                                                   & \textbf{F1}              & \textbf{Pr}              & \textbf{Re}              & \textbf{Acc}             \\ \hline
    \multicolumn{1}{c|}{\multirow{13}{*}{\rotatebox{90}{EyeQ}}}    & Wang et al. 2015 \cite{wang2015human}                      & 0.699          & 0.740          & 0.699          & -               \\ \cline{2-6} 
    \multicolumn{1}{c|}{}                          & Yan et al. 2018 \cite{yan2018two}                           & 0.748          & 0.798          & 0.745          & 0.793          \\ \cline{2-6} 
    \multicolumn{1}{c|}{}                          & Fu et al. 2019 \cite{fu2019evaluation}                     & 0.855          & 0.865          & 0.850          & -               \\ \cline{2-6} 
    \multicolumn{1}{c|}{}                          & Ou et al. 2019 \cite{ou2019novel}                          & 0.744          & 0.798          & 0.748          & 0.792          \\ \cline{2-6} 
    \multicolumn{1}{c|}{}                          & Raj et al. 2020 \cite{raj2020multivariate}                 & 0.869          & 0.870          & \textbf{0.970} & 0.884          \\ \cline{2-6} 
    \multicolumn{1}{c|}{}                          & Zhou et al. 2020 \cite{zhou2020retinal}                    & 0.868          & 0.866          & 0.870          & -               \\ \cline{2-6} 
    \multicolumn{1}{c|}{}                          & Muddamsetty \& Moeslund, 2021 \cite{muddamsetty2021multi}  & 0.859 & 0.864          & 0.857          & 0.881               \\ \cline{2-6} 
    \multicolumn{1}{c|}{}                          & Leonardo et al. 2022 \cite{leonardo2022impact}             & \textbf{0.878} & \textbf{0.879}         & 0.878          & 0.894          \\ \cline{2-6} 
    \multicolumn{1}{c|}{}                          & Xu et al. 2023 \cite{xu2023deep}                           & 0.872          & 0.876 & 0.871          & 0.889          \\
    \cline{2-6} 
    \multicolumn{1}{c|}{}                          & QuickQual \cite{engelmann2023quickqual}                          & 0.867          & 0.877 & 0.861          & 0.886          \\ \cline{2-6} 
    \multicolumn{1}{c|}{}                          & K\"onig et al. 2024 \cite{konig2024quality} & 0.830          & 0.810          & 0.850          & \textbf{0.910} \\ \cline{2-6} 
    \multicolumn{1}{c|}{}                          & Guo et al. 2024 \cite{Guo2024RefinedIQ} & 0.846          & 0.867          & 0.827          & 0.866 \\ \cline{2-6} 
    \multicolumn{1}{c|}{}                          & Huang et al. 2024\cite{huang2024enhancing} (DenseNet121) & 0.734          & 0.741          & 0.761          & 0.779 \\ \cline{2-6} 
    \multicolumn{1}{c|}{}                          & Huang et al. 2024\cite{huang2024enhancing} (SwinT-S) & 0.763          & 0.765          & 0.770          & 0.803 \\ \cline{2-6} 
    \multicolumn{1}{c|}{}                          & Huang et al. 2024 \cite{huang2024enhancing} & 0.776          & 0.791          & 0.765          & 0.815 \\ \cline{2-6} 
    \multicolumn{1}{c|}{}                          & ST-EyeQ                                               & 0.863          & 0.866          & 0.863          & 0.882          \\ \cline{2-6} 
    \multicolumn{1}{c|}{}                          & \textbf{MT-EyeQ (ours)}                                                  & \textbf{0.875} & \textbf{0.877} & \textbf{0.874} & \textbf{0.891} \\ \hline
    \multicolumn{1}{c|}{\multirow{3}{*}{\rotatebox{90}{\shortstack{Deep \\ DRID}}}}  & QuickQual \cite{engelmann2023quickqual}                                                                  & 0.742      & 0.698 & 0.790    & 0.698     \\ \cline{2-6}

    \multicolumn{1}{c|}{}& ST-DeepDRiD                                                        & 0.763          & \textbf{0.745}          & 0.782 & 0.732 \\ \cline{2-6} 
    
    \multicolumn{1}{c|}{}                          & \textbf{MT-DeepDRiD (ours)}                                         & \textbf{0.778} & 0.721 & \textbf{0.845}          & \textbf{0.735}  \\ \hline
    \end{tabular}}
    \label{tab:comparativa}
\end{table}

\begin{figure}[t]
    \centering 

    \begin{subfigure}{0.99\linewidth}
        \centering        \includegraphics[width=\linewidth]{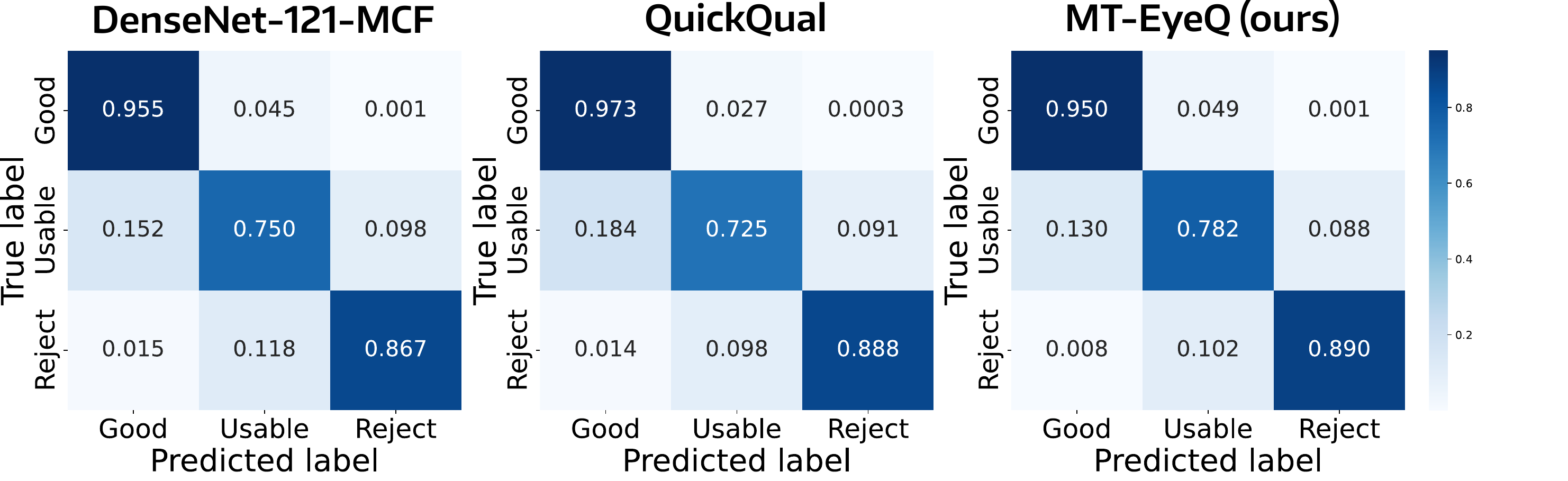} 
        \caption{EyeQ test set}
        \label{fig:imagen_a_m_confusion}
    \end{subfigure}
    \hfill 
    \vspace{0.5cm}
    \begin{subfigure}{0.60\linewidth}
        \centering        \includegraphics[width=\linewidth]{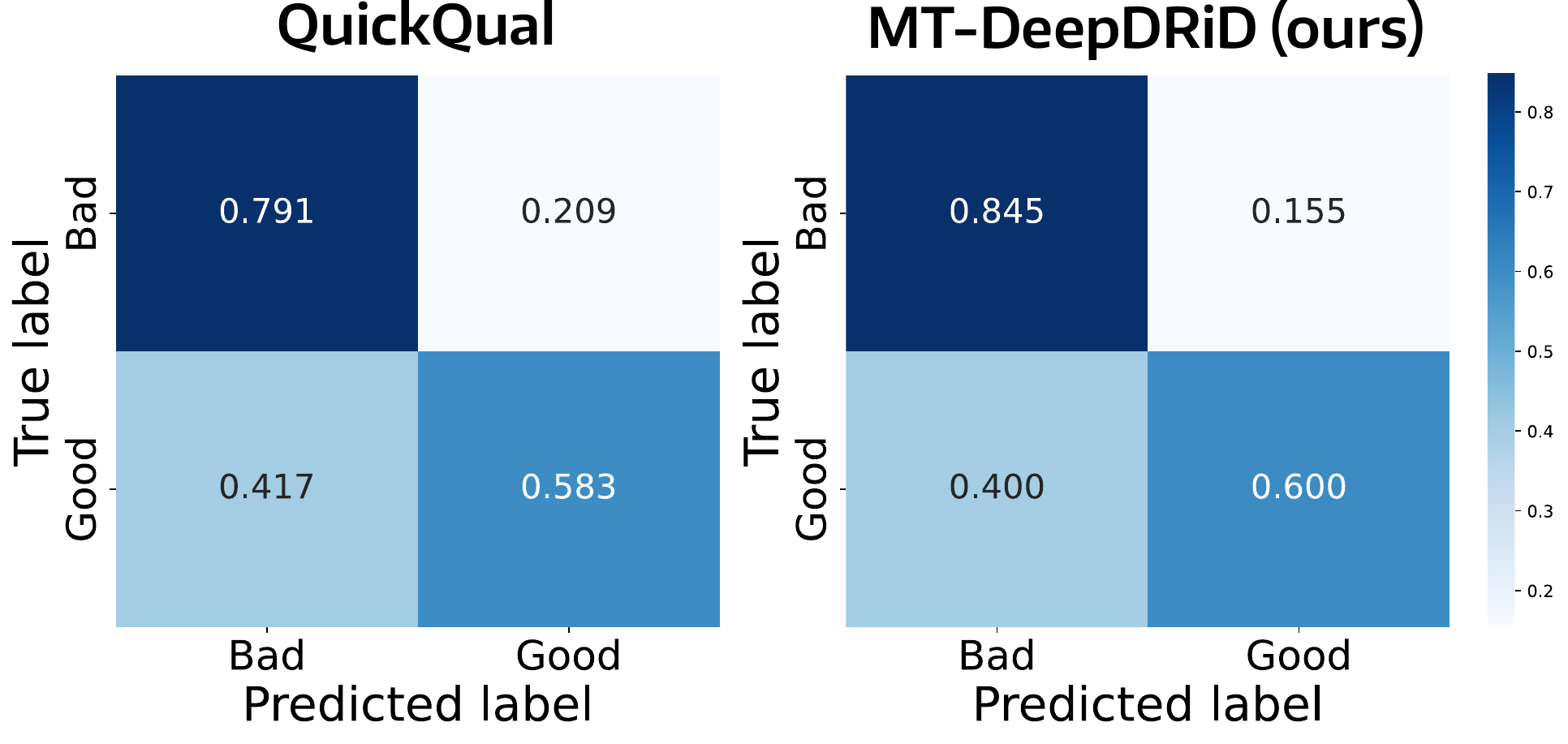}
        \caption{DeepDRiD test set}
        \label{fig:imagen_b_m_confusion}
    \end{subfigure}

    \caption{Per-row normalized confusion matrices for overall quality classification on EyeQ and DeepDRiD test sets, as reported in DenseNet-121-MCF \cite{fu2019evaluation}, QuickQual \cite{engelmann2023quickqual} and as obtained by our method.}
    \label{fig:imagenes_conjunto_m_confusion}
\end{figure}

To complement the previous evaluation, Figure~\ref{fig:imagenes_conjunto_m_confusion} presents per-row normalized confusion matrices for our multi-task models on the EyeQ and DeepDRiD test sets, including other two state-of-the-art models that either reported their matrices~\cite{fu2019evaluation} or shared their code~\cite{engelmann2023quickqual} In EyeQ, our approach notably improves results for the 'Usable' class. For the 'Good' class, all models exhibit strong performance, with the highest one being reported by QuickQual. Finally, our model achieved the highest accuracy for the 'Reject' class. In DeepDRiD, on the other hand, our model improved classification for both the 'Good' and 'Bad' classes compared to the re-trained QuickQual, which is consistent with the summary observed in Table~\ref{tab:comparativa}.

To examine per-class contributions to macro-F1 obtained in EyeQ, Table~\ref{tab:f1-per-class} reports one-vs-all F1 for ST-EyeQ and MT-EyeQ, and for the two competing models included in Figure~\ref{fig:imagenes_conjunto_m_confusion}. Gains are observed as consistent across classes, and are higher than those reported by other state-of-the-art approaches. For further reference, Yi et al.~\cite{yi2023label} modeled RIQA as a binary task (merging ``usable'' and ``reject'' into a bad-quality class) and reported an F1 of 0.895 for detecting the good-quality class. Similarly, Guo et al.~\cite{Guo2024RefinedIQ} achieved 0.926 for the same task. In contrast, our MT-EyeQ achieved 0.940 when evaluated using this setting. The largest improvement is seen for the ``Usable'' category, where F1 increases by more than 2 percentage points. This class is inherently ambiguous: images may contain artifacts yet remain suitable for diagnosis.

    

\begin{table}[t]
\centering
\caption{Comparison of per-class $F_1$-scores for overall quality classification on the EyeQ test set for the proposed multi-task model (MT-EyeQ) against the ST-EyeQ, MCF \cite{fu2019evaluation}, and QuickQual \cite{engelmann2023quickqual} models}
\label{tab:f1-per-class}
\resizebox{0.7\textwidth}{!}{ 
\begin{tabular}{c|ccc}
\hline
\multirow{2}{*}{\textbf{Model}} & \multicolumn{3}{c}{\textbf{F1-score}} \\ \cline{2-4} 
 & \textbf{Reject} & \textbf{Usable} & \textbf{Good} \\ \hline
MCF \cite{fu2019evaluation}  & 0.864 & 0.783 & 0.935 \\ \hline
QuickQual \cite{engelmann2023quickqual} & 0.880 & 0.786 & 0.937 \\ \hline
ST-EyeQ & 0.872 & 0.782 & 0.936 \\ \hline
\textbf{MT-EyeQ (ours)} & \textbf{0.883} & \textbf{0.804} & \textbf{0.940} \\ \hline
\end{tabular}
}
\end{table}

\begin{figure}[t!]
  \centering
    {\includegraphics[width=0.95\textwidth]{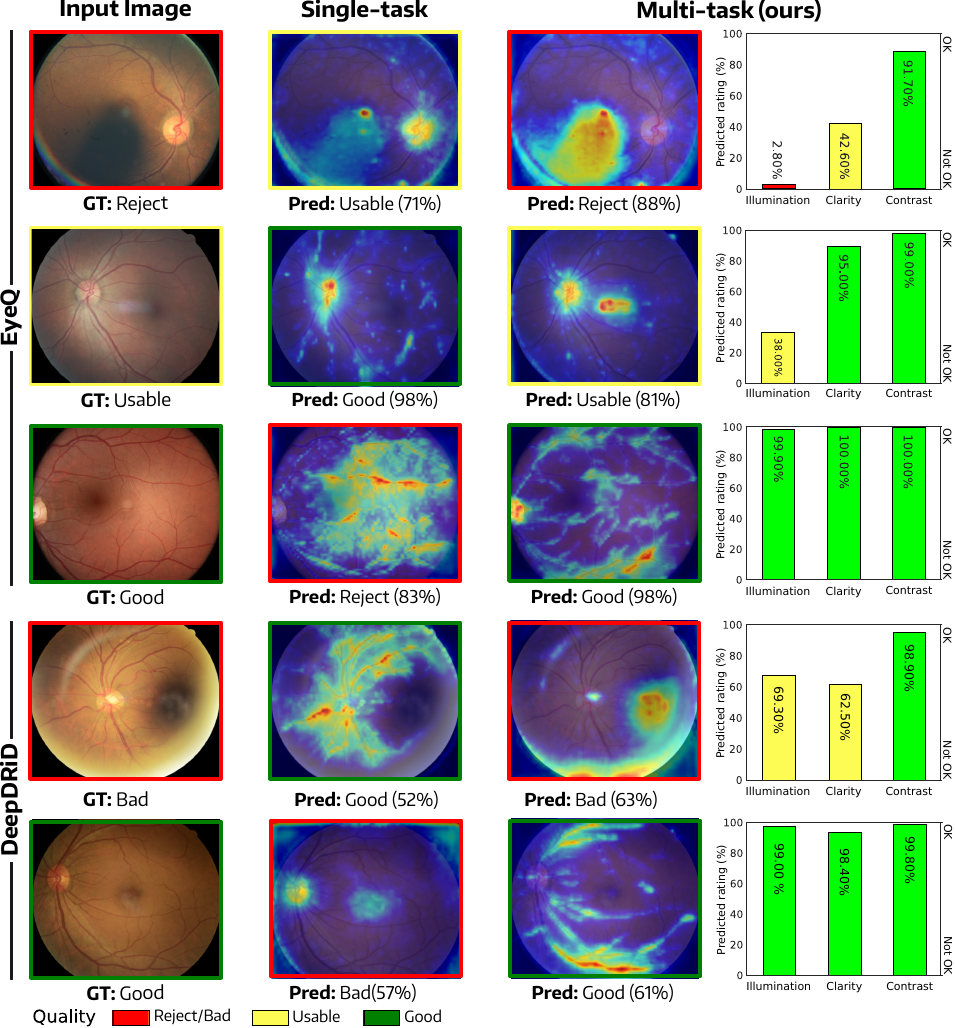}}
    {\caption{Qualitative comparison of results obtained in EyeQ (top) and DeepDRiD (bottom) by the single-task (left) and multi-task (right) models.}
    \label{fig:GradCams}}
\end{figure}

Figure~\ref{fig:GradCams} shows representative test images from EyeQ (top) and DeepDRiD (bottom)--one per overall-quality class (rejectable, usable, good for EyeQ; bad and good for DeepDRiD)--together with outputs and GradCAMs from the single- and multi-task models. From left to right, the figure depicts the input image and its ground truth (GT) label, the GradCAMs and class predictions from both approaches, and the quality-detail outputs (task $A$) from our multi-task model. All examples correspond to misclassification by the single-task model that were correct when using the multi-task variant.

The first EyeQ example is rejectable, with a shadow affecting the macular region, a rainbow artifact in the inferior temporal quadrant, flash artifacts in the superior nasal quadrant, and mild global blur. The single-task model predicted ``usable'' and activated GradCAM chiefly on the optic disc (likely as a proxy for usability), the fovea (possibly the flash spot), and the rainbow artifact. The multi-task model correctly predicted ``rejectable'' and highlighted the most problematic areas 

The second EyeQ case is ``usable'' with a flash artifact partially covering the fovea. The single-task model predicted ``good'' with high confidence, focusing its GradCAM on the disc and vascular structures. The multi-task model matched the ground truth and highlighted both the disc and the flash artifact.

The third EyeQ image, labeled ``good,'' shows excellent visibility of retinal structures, although part of the central retinal vein and the optic disc fall outside the FOV. The single-task model predicted ``bad,'' with a diffuse GradCAM that is hard to interpret. The multi-task model correctly predicted ``good'' with high confidence, indicated good illumination, clarity, and contrast, and emphasized the partially visible disc and vascular arcades.

The final two examples come from DeepDRiD. The first is a bad-quality, disc-centered image that the single-task model predicted as ``good'' (with high uncertainty); its GradCAM emphasizes the vascular arcades and the disc. The multi-task model predicted ``bad'' and concentrated GradCAM on problematic regions such as the FOV edge and the macular area, where strong illumination artifacts obscure the fovea. The last example is a good-quality image misclassified by the single-task model as ``bad,'' with GradCAM activations scattered even outside the FOV. The multi-task model correctly predicted ``good'' and emphasized the visibility of key structures, including the vascular arcades.

Figure~\ref{fig:GradCams} also shows that ST-EyeQ made some errors with high confidence (e.g., predicting a good-quality image as rejectable with 83\% probability). With MT-EyeQ, predictions shift toward the correct class, again with high confidence. In DeepDRiD, ST-DeepDRiD’s errors occur with lower confidence, and MT-DeepDRiD’s correct predictions are less confident than those in EyeQ.

\begin{figure}[t!]
  \centering
    {\includegraphics[width=0.99\textwidth]{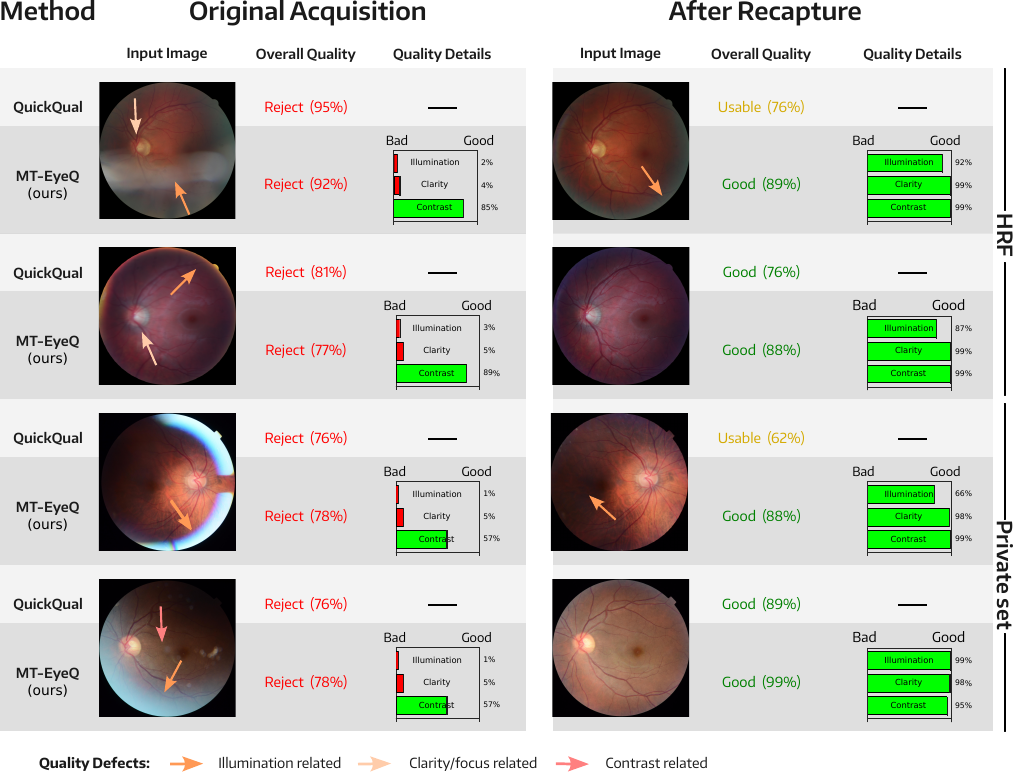}}
    {\caption{Comparison of our MT-EyeQ model against QuickQual \cite{engelmann2023quickqual} in image recapture scenarios. Pairs of initial low-quality captures are presented alongside their improved versions from the HRF \cite{odstrcilik2013retinal} and private datasets.}}
    \label{fig:CasosUso}
\end{figure}

To illustrate the clinical applicability of our approach, Figure~\ref{fig:CasosUso} compares the MT-EyeQ model with the QuickQual~\cite{engelmann2023quickqual} baseline in real-world image recapture scenarios, using examples from HRF~\cite{odstrcilik2013retinal} and a private dataset. For the original low-quality acquisitions, both models correctly classified the images as “Reject.” However, MT-EyeQ additionally provides interpretable quality details (e.g., illumination, clarity, and contrast issues) that can guide technicians during recapture.

In the first case, the arrows highlight illumination defects and blurring in the optic disc area. MT-EyeQ identified insufficient illumination and clarity, which improved after recapture, though the model still detected minor illumination artifacts (i.e. the bright ring due to an insufficiently dilated pupil). In the second case, bad illumination (likely due to the bright artifact in the superior temporal and nasal quadrants) and presence of blurring (i.e. in the optic disc area) were again flagged by the MT-EyeQ model, and the recaptured image showed better quality, albeit slightly dark, as noted by our model. In the third case, illumination artifacts were clearly visible, with our model detecting it correctly but also reporting poor clarity; QuickQual classified the recaptured image as “Usable”, while MT-EyeQ correctly labeled it as good quality, indicating however some residual illumination issues (i.e. due to poor visibility of the macular area). Finally, in the fourth case, beyond identifying poor quality, our model’s capture-condition branch discriminated inadequate illumination and clarity and relatively poor contrast, all of which were improved and corrected after recapture.

\subsection{Quality details}
\label{subsec:results:capture}

To evaluate the semi-supervised quality-detail task, we assessed MT-EyeQ and MT-DeepDRiD on MSHF and EyeQ-D, which provide ground-truth labels for this task (see Section~\ref{subsec:materials}). We also compared both models with the Teacher-A used to pseudo-label their training sets to test whether the Students retain the Teacher’s accuracy. Table~\ref{tab:detail-evaluation-mshf-eyeqd} summarizes results in terms of F1, Pr, and Re.

Overall, both multi-task models achieve F1 values that are statistically comparable to Teacher-A ($p>0.05$) across datasets. In MSHF, MT-EyeQ shows a significant difference only for contrast ($p<0.05$), with a lower F1 than Teacher-A (0.875 vs.\ 0.940). In EyeQ-D, F1 is higher for illumination than Teacher-A, driven by a small increase in Re, but no task shows a statistically significant difference ($p>0.05$). In MSHF, MT-DeepDRiD attains a higher F1 for clarity than both Teacher-A and MT-EyeQ--primarily due to improved Re--though differences are not significant ($p>0.05$). For illumination and contrast, its performance is lower with significant differences ($p<0.05$). A similar drop for illumination and contrast appears in EyeQ-D ($p<0.05$). Across both datasets, MT-EyeQ outperforms MT-DeepDRiD.

\begin{table}[t!]
  \caption{Evaluation of quality detail prediction on the MSHF and EyeQ-D datasets. The * symbol in F1 denotes statistically significant differences compared to the Teacher-A model.}
  \label{tab:detail-evaluation-mshf-eyeqd}
  \resizebox{0.99\textwidth}{!}{

        \begin{tabular}{c| c | c | c | c | c | c | c | c | c | c}
        \hline
        \multirow{2}{*}{\textbf{Dataset}} & \multirow{2}{*}{\textbf{Model}} & \multicolumn{3}{c|}{\textbf{Illumination}}                                        & \multicolumn{3}{c|}{\textbf{Clarity}}                                             & \multicolumn{3}{c}{\textbf{Contrast}}                                             \\ 
        \cline{3-11} 
                                          &                                 & {\textbf{F1}} & {\textbf{Pr}} & \textbf{Re} & {\textbf{F1}} & {\textbf{Pr}} & \textbf{Re} & {\textbf{F1}} & {\textbf{Pr}} & \textbf{Re} \\ 
        \hline
        \multirow{2}{*}{MSHF \cite{jin2023mshf}}             & Teacher-A
        & {0.908}      & {0.864}      & 0.957      & {0.807}     & {0.909}      & 0.725      & {0.875}     & {0.837}      & 0.917      \\ \cline{2-11} 
                                                             & MT-EyeQ
       & {0.897}      & {0.902}      & 0.893      & {0.788}      & {0.825}      & 0.754      & {0.840*}      & {0.872}       & 0.810      \\ \cline{2-11}
                                                                & MT-DeepDRiD
        & {0.795*}      & {0.660}      & 1.000      & {0.844}      & {0.795}      & 0.899      & {0.812*}      & {0.695}      & 0.976      \\ \hline
        
        \multirow{2}{*}{EyeQ-D }           & Teacher-A
        & {0.755}      & {0.633}      & 0.934      & {0.919}      & {0.942}      & 0.897      & {0.881}     & {0.787}      & 1.000      \\ \cline{2-11} 
                                                                & MT-EyeQ
        & {0.758}      & {0.630}      & 0.951      & {0.921}      & {0.973}      & 0.873      & {0.876}      & {0.785}      & 0.991      \\ \cline{2-11} 
                                                                & MT-DeepDRiD
        & {0.670*}      & {0.504}      & 1.000      & {0.912}      & {0.965}      & 0.865      & {0.868*}      & {0.774}      & 0.991      \\ \hline
        \end{tabular}
    }
\end{table}

\begin{figure}[t!]
  \centering
    {\includegraphics[width=0.7\textwidth]{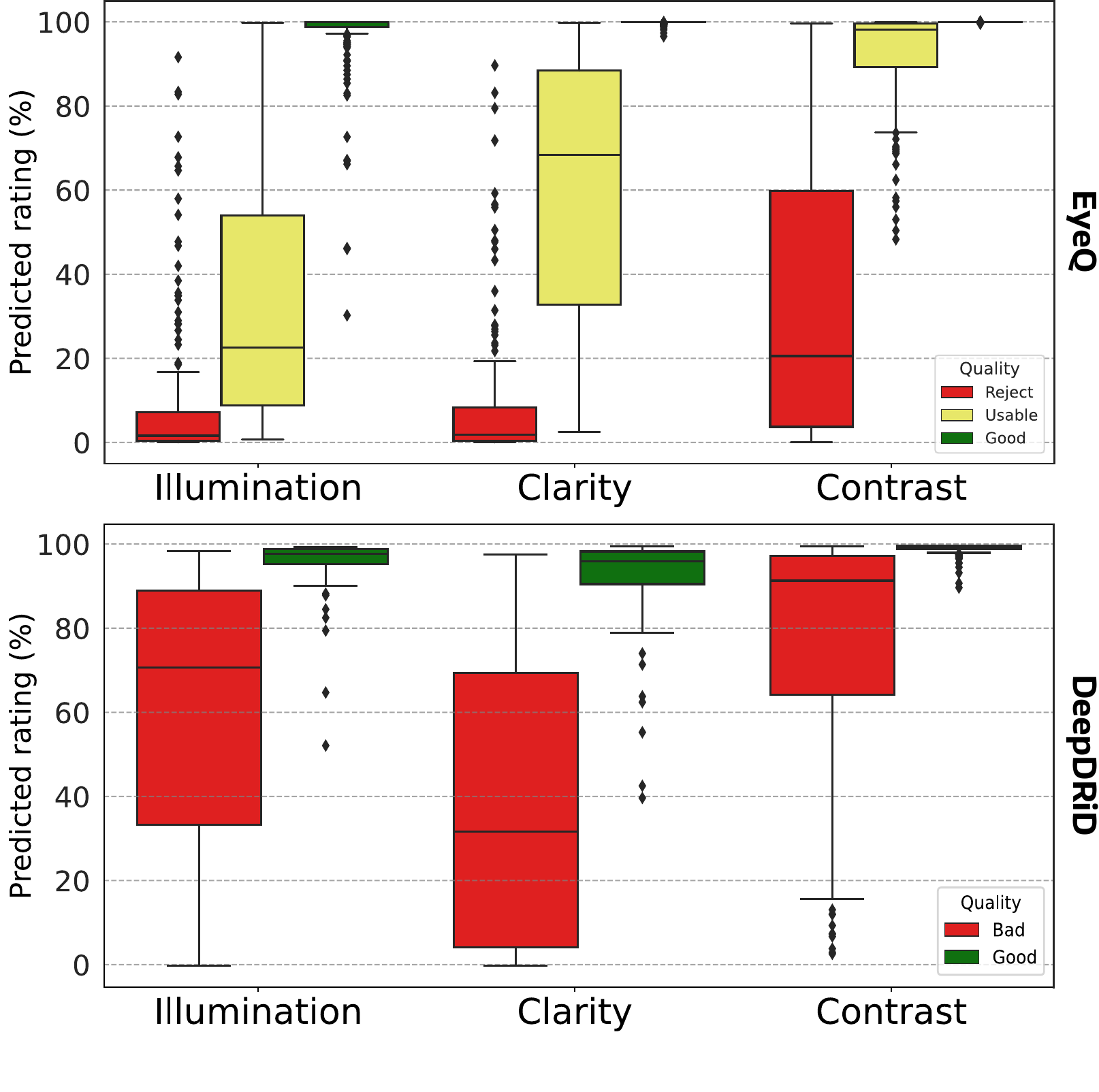}}
    {\caption{Distribution of ratings of good/bad illumination, clarity and contrast as predicted by our multi-task model in the EyeQ and DeepDRiD test set, for each ground truth overall quality group.}
    \label{fig:boxplot-class}}
\end{figure}

To complement these results, Figure~\ref{fig:boxplot-class} shows the distributions of predicted probabilities for the three quality details produced by MT-EyeQ and MT-DeepDRiD on their respective test sets, stratified by the ground-truth overall-quality label. In EyeQ, rejectable images receive low illumination and clarity scores from MT-EyeQ; contrast is more dispersed but with a low median. Usable images show higher scores for illumination, clarity, and contrast than rejectable images, with greater variability. Good-quality images cluster at high probabilities. Occasional high/low outliers in illumination and clarity for the rejectable and good-quality groups may reflect inconsistencies in manual overall-quality labels or errors in quality-detail predictions. In DeepDRiD, bad-quality images exhibit greater variability than in EyeQ across illumination, clarity, and contrast, plausibly due to the absence of an intermediate “usable” class that would separate borderline cases. Good-quality images show higher and more consistent scores across all details.

To relate pseudo-label noise to expert variability, we compared Teacher-A predictions with those of a single human evaluator. Specifically, we selected one ophthalmologist from EyeQ-D and formed a new ground-truth by majority vote among the remaining experts (see Section~\ref{subsec:materials}). The chosen evaluator showed the highest intra-rater consistency when, blinded to prior labels, re-annotating previously seen images. Table~\ref{tab:noise-vs-experts} presents the comparison. Note that Teacher-A results differ slightly from Table~\ref{tab:detail-evaluation-mshf-eyeqd} because the ground truth changes when one expert is held out. Most metrics are consistent between Teacher-A and the expert, with no statistically significant differences in F1 ($p>0.05$). The only significant gaps are in Pr and Re for illumination and contrast ($p<0.05$).

Figure~\ref{fig:GradCams} also illustrates quality-detail predictions from the multi-task models. In the rejectable EyeQ example, the model flags illumination issues and partial clarity defects consistent with the input (shadows, flash artifacts, and blur) and highlights these regions in the GradCAM (see Section~\ref{subsec:results:quality}). A similar pattern appears in the bad-quality DeepDRiD example, where illumination and clarity issues are present but less pronounced. For the EyeQ “usable” case, the model indicates poor illumination—most likely due to the light artifact over the fovea. Finally, both good-quality examples in EyeQ and DeepDRiD are predicted as well illuminated with good clarity and contrast.

\begin{figure}[t!]
    \centering 

    \begin{subfigure}{0.99\textwidth}
        \centering
        \includegraphics[width=0.99\textwidth]{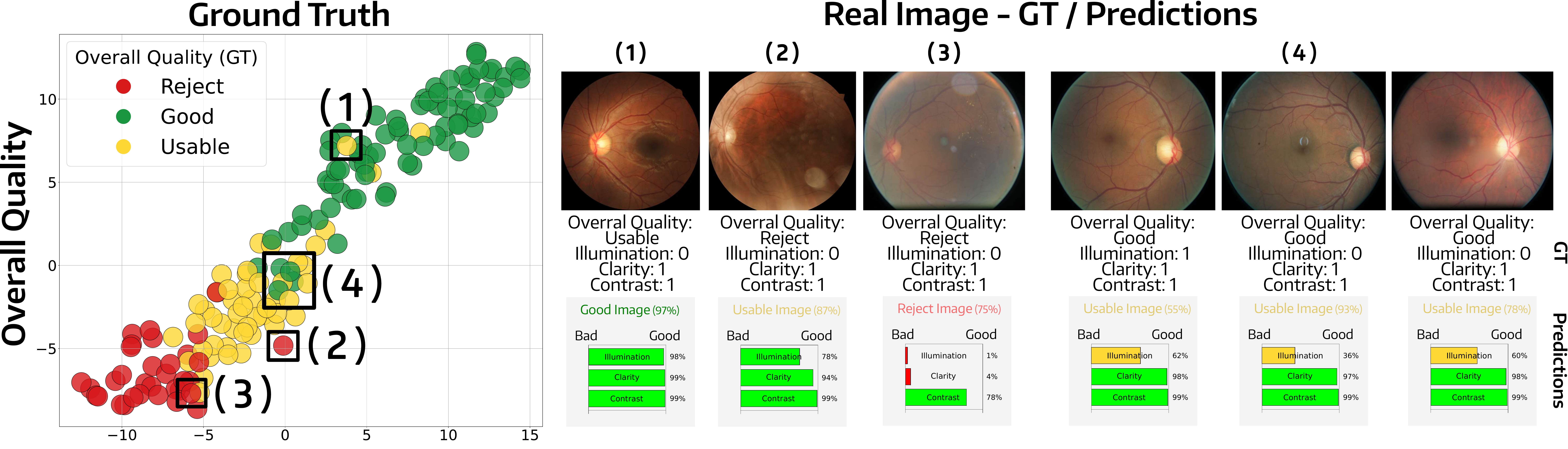}
        \caption{} 
        \label{fig:plot-tsne-quality-miniature}
    \end{subfigure}
    
    \vspace{0.25cm} 
    
    \begin{subfigure}{0.99\textwidth}
        \centering
        \includegraphics[width=0.99\textwidth]{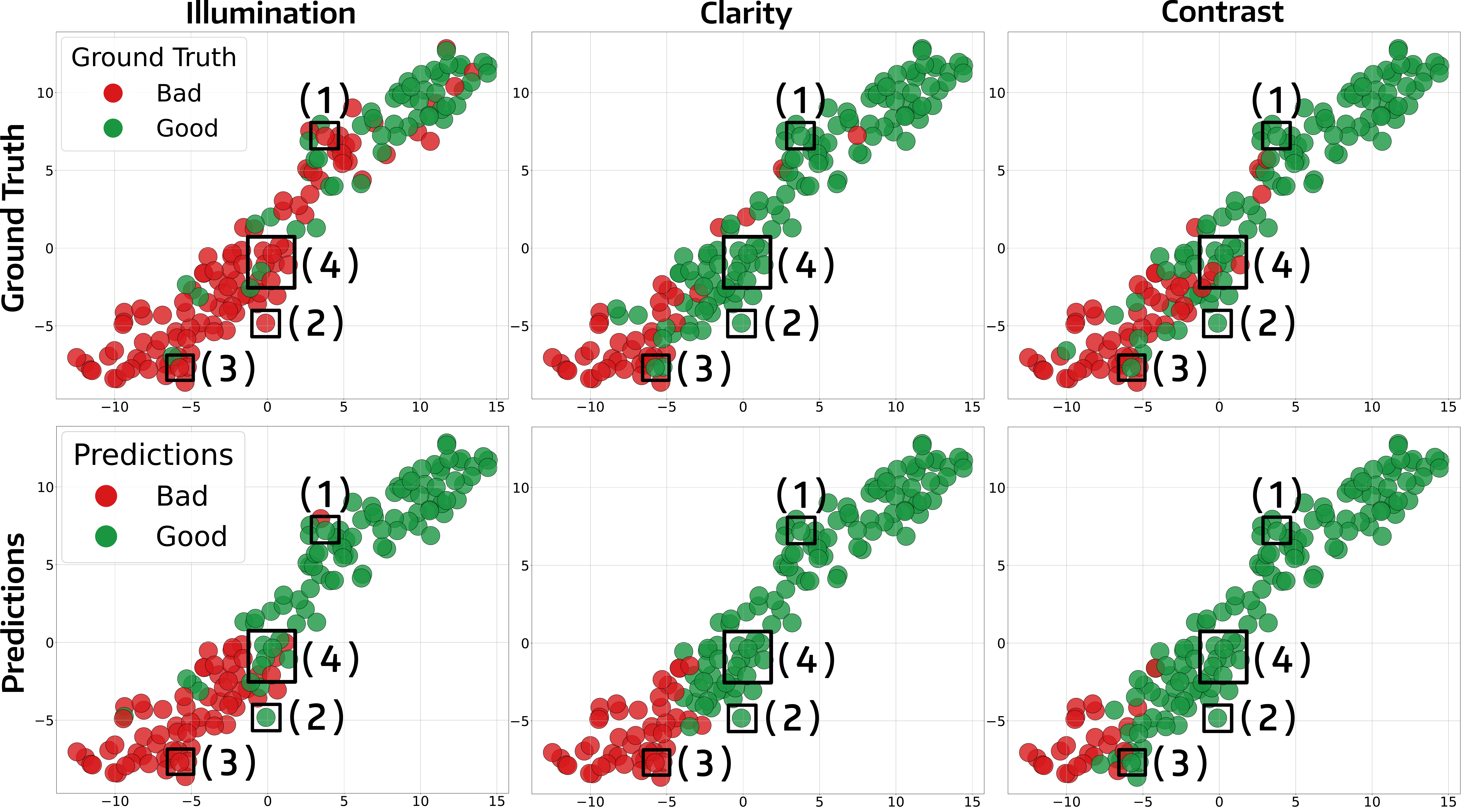}
        \caption{}
        \label{fig:imagen_b}
    \end{subfigure}

    \caption{t-SNE visualization of the feature space on the EyeQ-D test set. (a) Features colored by ground-truth overall quality labels, with specific examples (1-4) highlighted. (b) Features colored by quality detail labels, using ground truth (top) and predictions (bottom).}
    \label{fig:imagenes_conjunto_plots_tsne}
\end{figure}

To understand the relationship in the latent space between overall quality and quality details, Figure~\ref{fig:imagenes_conjunto_plots_tsne} depicts the t-SNE projection of the learned representations of images in EyeQ-D, which is shared by both tasks.

\begin{table}[t!]
    \caption{Comparison of the Teacher-A model and a human expert on quality detail prediction in the EyeQ-D subset. The * symbol denotes statistically significant differences with respect to the counterpart.}
  \label{tab:noise-vs-experts}
  \resizebox{0.99\textwidth}{!}{

        \begin{tabular}{c|ccc|ccc|ccc}
        \hline
        \multirow{2}{*}{\shortstack{\textbf{Evaluated} \\\textbf{expert / model}}} & \multicolumn{3}{c|}{\textbf{Illumination}}                                           & \multicolumn{3}{c|}{\textbf{Clarity}}                                                & \multicolumn{3}{c}{\textbf{Contrast}}                                               \\ \cline{2-10} 
                                   & \multicolumn{1}{c|}{\textbf{F1}}              & \multicolumn{1}{c|}{\textbf{Pr}}     & \textbf{Re}     & \multicolumn{1}{c|}{\textbf{F1}}              & \multicolumn{1}{c|}{\textbf{Pr}}     & \textbf{Re}     & \multicolumn{1}{c|}{\textbf{F1}}              & \multicolumn{1}{c|}{\textbf{Pr}}     & \textbf{Re}     \\ \hline
        Ophthalmologist                   & \multicolumn{1}{c|}{0.707}          & \multicolumn{1}{c|}{0.552}    & 0.982*    & \multicolumn{1}{c|}{0.934} & \multicolumn{1}{c|}{0.941} & 0.926 & \multicolumn{1}{c|}{0.871}          & \multicolumn{1}{c|}{0.817*} & 0.934 \\ \hline
        Teacher-A              & \multicolumn{1}{c|}{0.708} & \multicolumn{1}{c|}{0.567*} & 0.944 & \multicolumn{1}{c|}{0.930}          & \multicolumn{1}{c|}{0.934} & 0.926 & \multicolumn{1}{c|}{0.871} & \multicolumn{1}{c|}{0.772} & 1.0*    \\ \hline
        \end{tabular}
    }
\end{table}

Figure~\ref{fig:imagenes_conjunto_plots_tsne}(a) (left) is colored using the ground truth labeling. Three well-defined clusters are observed, one for each of the overall quality classes, with transitions between groups that are consistent with the semantics of the classes (i.e. from 'Good' to 'Usable' and then 'Reject').

Figure~\ref{fig:imagenes_conjunto_plots_tsne}(b) includes the same t-SNE representation used in (a), but colored by the ground truth good/bad labeling of quality details (top) and model prediction (bottom). In general, the transition between good and bad illumination, clarity and contrast is consistent with respect to the overall quality categories depicted in Figure~\ref{fig:imagenes_conjunto_plots_tsne}(a). When comparing predictions against ground truth labels, illumination is less consistent, with experts labeling fewer images as well-illuminated than the proposed approach. This situation is different for clarity and contrast, in which both groups are relatively similar, with the exception of a small group of low contrast images located at the transition to the good contrast images.

Errors and special cases are shown on the right-hand side of Figure~\ref{fig:imagenes_conjunto_plots_tsne}(a) and are correspondingly marked on the t-SNE maps. Image (1), manually labeled as “Usable,” was misclassified by the model as “Good” and clustered among images labeled as such. The model predicted all quality details as good, although experts noted poor illumination--likely because the model emphasizes illumination artifacts rather than overall darkness.
Image (2) shows a similar case: it was labeled “Bad” but predicted as “Usable.” Although the model flagged suboptimal illumination, the probability remained above 50\%, preventing a “poor illumination” classification.
Image (3), a “Reject” case, was correctly classified; however, the model also predicted poor clarity, whereas experts rated clarity as good--possibly because the illumination artifact degraded the focus perceived by the model.
Finally, (4) depicts a group of three high-quality samples located at the boundary between the “Good” and “Usable” clusters, all classified as “Usable” by the model, with varying levels of certainty. The first case was classified as “Usable” with a 55\% probability. When analyzing quality details, the model predicted all correctly, but with slightly low illumination, possibly related to peripheral artifacts from insufficient pupil dilation. The second case was classified as “Usable” with high certainty. Experts agreed that the image had poor illumination (i.e., due to the bright artifact over the macular region), and the model correctly captured this characteristic. Finally, the third case was predicted as “Usable” with higher certainty than the first case. It was flagged as poorly illuminated by the readers (i.e., due to the shadow in the inferior quadrants), and while the model did not classify it as poorly illuminated, it predicted good illumination with 60\% certainty.

\section{Discussion}
\label{sec:discussion}

We propose a multi-task learning framework that leverages pseudo-labels generated by a model trained on limited data to develop more interpretable RIQA models than standard single-task approaches. We hypothesize that incorporating an auxiliary task closely related to the overall quality assessment objective can enhance the original model's performance, even when labels for the auxiliary task are automatically generated by another model.

A key advantage of our multi-task learning approach is its ability to enhance model interpretability without requiring costly, large-scale data annotation for additional supervision. 

Traditional single-task RIQA models often become biased during training, focusing primarily on the integrity of anatomical structures—such as the optic disc and vascular arcades—since these regions are indeed correlated with good overall image quality. This bias can be observed in Figure~\ref{fig:GradCams}, where the usable case from EyeQ and the bad case from DeepDRiD were incorrectly classified as good quality by the single-task model, likely because both the optic disc and vessels were visible (as indicated by the corresponding GradCAMs). The opposite behavior is observed in the good-quality example from EyeQ, also in Figure~\ref{fig:GradCams}, which was incorrectly classified as bad quality by the single-task model. In this case, both the optic disc and the central retinal vein are partially visible or absent due to the restricted field of view (FOV). 
However, image quality is not solely determined by the visibility of these anatomical structures. As observed in both examples, despite the presence of vessels and the optic disc, experts also consider other critical factors, such as overall illumination, focus, the absence of artifacts, and contrast. These characteristics are crucial for the clinical utility of retinal images, particularly in diabetic retinopathy (DR) assessment, where detecting microaneurysms, small hemorrhages, or early ischemic areas is essential for identifying referable cases.

By fine-tuning a single-task model in a multi-task setting with an auxiliary task that predicts quality details, the model is pushed to focus more in actual acquisition defects, correcting these classification errors. This is exposed by the gains in performance reported in Table~\ref{tab:comparativa}, in which our multi-task approach consistently outperforms its single-task counterparts. 
As observed in Figure~\ref{fig:GradCams}, the multi-task models not only correct previous mistakes but also gain more certainty for the right class, predicting it with a high probability. 
Furthermore, these changes make GradCAMs associated to this model emphasize more evidently on the problematic areas, such as overexposed macular regions, blurry vessel structures, or field-of-view artifacts—features that are directly relevant to image quality assessment. Furthermore, when applied to good-quality scans, the GradCAMs continue to exhibit high activations in visible anatomical areas, ensuring that essential retinal structures remain highlighted (Figure~\ref{fig:GradCams}).

The idea of combining overall quality assessment with the categorization of quality details is not new and has been previously explored by other authors \cite{shen2018multi, shen2020domain, konig2024quality}. However, these models are typically trained on limited datasets, leading to poor domain generalization when evaluated on external datasets \cite{konig2024quality}. This limitation primarily arises from the fact that their training sets are manually labeled for both tasks. The fine granularity of quality detail labels makes the expert-reliant manual annotation process costly and significantly time-consuming. This is because aspects such as illumination, contrast, and sharpness are inherently subjective and difficult to standardize, introducing variability in the labels even among experienced ophthalmologists. As shown in Table~\ref{tab:noise-vs-experts}, even the most consistent annotator in our study exhibits a noise level statistically comparable to that of the Teacher model when performing this task.

Consequently, while existing multi-task approaches tend to be more accurate, they are also more expensive and less scalable than their single-task counterparts, hindering their real-world adoption. To address this limitation, we propose training a Teacher model on a small manually labeled dataset (e.g., MSHF), and then use it to generate pseudo-labels for larger training sets with available labels for overall quality. By relying on fewer samples for learning the Teacher (only 642 vs. K\"onig et al. approach~\cite{konig2024quality}, which uses 1922), we are inherently reducing the annotation burden and automating the labeling of this additional task in the larger set, in a semi-supervised manner. Our results indicate that, despite the use of pseudo-labels, the multi-task model still achieves quality detail predictions that are statistically comparable to those of the Teacher model (Table~\ref{tab:detail-evaluation-mshf-eyeqd}) and the interobserver variability (Table~\ref{tab:noise-vs-experts}).

We believe this occurs because the interobserver variability (i.e., the intrinsic noise present in expert-generated labels) is comparable to the errors introduced by the Teacher model when pseudo-labeling larger datasets (Table~\ref{tab:noise-vs-experts}). As a result, fine-tuning a pre-trained model for overall quality assessment on a sufficiently big dataset helps to partially absorb this noise, allowing the model to extract useful information from pseudo-labeled quality details to improve overall quality classification. This effect is evident in the performance improvements observed in Table~\ref{tab:comparativa}, where both multi-task models outperform their single-task counterparts, and also in the class-level comparisons in both Table \ref{tab:f1-per-class} and Figure \ref{fig:imagenes_conjunto_m_confusion}. Notably, our model reduces confusion for the ambiguous 'Usable' class, which is the most difficult to classify due to its ambiguity, at the cost of marginally loosing accuracy for the 'Good' class. In general, we observed this is caused by borderline 'Good' images flagged now as 'Usable'  (i.e. the cases depicted in Fig.~\ref{fig:plot-tsne-quality-miniature}). 
Moreover, the performance differences observed between MT-EyeQ and MT-DeepDRiD in Tables~\ref{tab:comparativa} and \ref{tab:detail-evaluation-mshf-eyeqd}, trained with 12543 images and 1200 images, respectively, further reinforce the importance of expanding the training set as much as possible to enhance model robustness.

The improvement in general quality classification can be attributed to the close relationship between this task and the assessment of acquisition details that influences quality. This relationship is evident in Figure~\ref{fig:boxplot-class}, where lower-quality images consistently exhibit reduced probabilities of good illumination, clarity, and contrast. Moreover, this relationship follows specific patterns depending on the granularity of the labels, as observed when comparing our results in EyeQ and DeepDRiD, two sets that define quality in three and two categories, respectively. Our experiments indicate that incorporating the auxiliary task improved the performance of the single-task model in both datasets, reinforcing the effectiveness of the multitask approach. In particular, in EyeQ, this improvement was observed across all three general quality categories, as reported in Table~\ref{tab:f1-per-class}, with the largest increase in the Usable class (2\%). This result is particularly relevant given that this category is the most ambiguous as it represents the boundary class, featuring images with good-quality characteristics but minor alterations that could be improved.
To better understand the task relationship within the multi-task core's feature space, Figure~\ref{fig:imagenes_conjunto_plots_tsne} shows that the clusters formed by overall quality (Figure~\ref{fig:imagenes_conjunto_plots_tsne}a) closely align with those for capture details (Figure~\ref{fig:imagenes_conjunto_plots_tsne}b), proving that the model learns a cohesive representation. Interestingly, Figure~\ref{fig:imagenes_conjunto_plots_tsne}b also highlights that illumination remains being a challenging detail to predict, with a less accurate point cloud with respect to the ground truth. This is consistent with the findings in Table~\ref{tab:noise-vs-experts}, where experts also struggled to agree. This might require using more granular labels for this characteristic, i.e. distinguishing by specific defects such as rainbow artifacts, peripheral reflections due to poorly dilated pupils, dark scans, etc.

Our approach outperforms most of the methods trained and evaluated on EyeQ (Table~\ref{tab:comparativa}), including state-of-the-art techniques based on Vision Transformers~\cite{dosovitskiy2020image}, such as the specialized Swin Transformer~\cite{liu2021swin} trained by Huang et al. \cite{huang2024enhancing} for quality assessment. We argue that for these two, the main problem is related with the lack of enough training samples, which has been reported to hamper the applicability of Transformers due to their overfitting tendency~\cite{dosovitskiy2020image}. 
Furthermore, notice that data augmentation needs to be controlled when training quality assessment models, as too aggressive transformations might generate samples that are contradictory with their associated label~\cite{leonardo2022impact}. This limits the possibility to artificially increase the training set with too much augmentation to reduce overfitting in Transformers.
Alternatively, we proposed leveraging a convolutional backbone such as a ResNet-18~\cite{he2016deep}, whose inductive biases--i.e. locality and translation equivariance--mitigate data needs while proving effective for our specific goal. Moreover, we used an augmentation strategy inspired in RandAugment~\cite{cubuk2020randaugment} that we tailored specifically for this application, with transformations limited to valid ranges that were experimentally determined (Section~\ref{subsec:experimental-setup-model-baseline}). 
Nevertheless, notice that our method is straightforward enough to be applied with bigger and more robust backbone architectures, as it only requires training for the objective $B$, incorporating an additional classification head for task $A$, and fine-tuning on the combined manual and pseudo-labels.

Compared to recent RIQA approaches with similar objectives, our method offers a favorable efficiency trade-off between interpretability, accuracy, and annotation cost. For instance, König et al.\cite{konig2024quality} employed a multi-label approach to predict both overall quality and quality details, but relied entirely on fully annotated datasets, limiting scalability and generalization capacity. Alternatively, our approach offered additional explainability at a fraction of their annotation cost, while also ensuring better performance (Table~\ref{tab:comparativa}). On the contrary, Leonardo et al.~\cite{leonardo2022impact} achieved a slightly higher overall quality prediction with its EfficientNet-B0 regression-based model, but without providing task-specific feedback on acquisition conditions. Our model reports a performance that is slightly lower, but incorporates a set of non-negligible explainable outputs that improves the usability of the model. Figure~\ref{fig:CasosUso} shows an exemplary use case scenario. Unlike the state-of-the-art approaches that only determine if an image should be rejected, our MT-EyeQ model complements this prediction with interpretable quality details that are effectively and accurately captured, and can actively guide the operator to recapture the scan.

Notice also that our approach is able to achieve high performance and interpretability even though we adopted a lightweight backbone architecture such as a ResNet-18. We chose such a small capacity network to reduce the risk of overfitting, a common issue i.e. with Transformer-based methods~\cite{dosovitskiy2020image}.This network outperformed many other more complex architectures with more parameters than ours (Table~\ref{tab:comparativa}), including Transformers and CNNs. This reflects that the proposed multi-task formulation can efficiently steer the model towards learning more informative and clinically relevant representations without requiring drastic changes in architecture nor largely increasing its capacity.

In terms of limitations of this study, the primary one is the narrow exploration of model choices. We deliberately used a compact backbone (ResNet-18) and a classification objective for overall quality to isolate the contribution of the semi-supervised multi-task design and to reduce overfitting risk given the modest data available to train the Teacher. We did not evaluate deeper or larger encoders, nor regression objectives as in \cite{leonardo2022impact}; we believe both alternatives could eventually lift absolute performance while preserving the proposed training scheme. Future extensions of this work could include adopting stronger encoders or foundation models pre-trained self-supervised on unlabeled fundus images and revisiting the overall-quality head as a regression task, which may improve accuracy without extra manual labels. A second limitation is that our results in EyeQ lie close to--but do not surpass--the best published figures. However, it is worth noting that our model supplies an additional, actionable output in the form of capture-attribute predictions; the previous extensions could plausibly close this gap. A third limitation is the evident dependence on the Teacher producing pseudo-labels--student performance remains bounded by the quality of this supervision--suggesting benefits from stronger pretraining for the Teacher. Finally, evaluation of quality-detail predictions relied on two relatively small sets (MSHF and the 160-image EyeQ-D subset labeled by eight ophthalmologists), which limits statistical power despite careful annotation; we release EyeQ-D to support replication and broader comparison, but future extensions in this regard could benefit achieve better statistical analyses.

\section{Conclusions}
\label{sec:conclusions}

We present a semi-supervised, multi-task framework for RIQA that uses pseudo-labels of capture attributes (illumination, contrast, clarity) learned from a small expert-annotated set to improve overall quality classification and provide immediate, actionable guidance for image recapture. Empirically, we demonstrated that pseudo-labels with noise comparable to inter-expert variability function as effective auxiliary signals, regularizing multi-task training and enhancing representation quality and interpretability while limiting annotation cost. This approach offers a step toward reducing the cost of building more explainable multi-task RIQA models and, more broadly, potentially other medical imaging systems. The main shortcomings of the proposed approach are its reliance on a Teacher model trained on modest training data, which limits the application of other architectures with more capacity. Future work could explore mitigating the overfitting risk using larger models pre-trained with self-supervised learning on unlabeled fundus images, aiding to improve Teacher's generalization ability without requiring additional training labels. Furthermore, we recommend exploring other problem formulations such as regression objectives, and integrating complementary pseudo-labels such as coarse anatomical cues, which can steer the model to explicitly look for those regions.

\subsubsection*{Acknowledgments}
This study was partially funded by UNICEN's 03-JOVIN-37c, CONICET's PIP GI 2021-2023 11220200102472CO, and Agencia I+D+i's PICTs 2019-00070 and 2021-00023. We also thank NVIDIA Corporation for granting 500 hours of GPU computation through a NVIDIA Applied Research Accelerator Program.

\subsubsection*{Statement of ethics}
This study does not involve human participants, animals, or identifiable personal data. All datasets
used in this research are publicly available and have been previously published by their respective
sources. Therefore, no ethical approval was required, and there are no ethical conflicts to disclose.

\subsubsection*{Declaration of Competing Interest}
The authors declare that they have no known competing financial interests or personal relationships
that could have appeared to influence the work reported in this paper.

\bibliographystyle{elsarticle-num}
\bibliography{samplebibliography}





\end{document}